\documentclass[runningheads]{llncs}

\usepackage[T1]{fontenc}
\usepackage{graphicx}
\usepackage{booktabs}
\usepackage{amsmath}
\usepackage{multirow}
\usepackage[table]{xcolor}
\usepackage{fvextra} 
\usepackage[hidelinks]{hyperref}

\newcommand{\best}[1]{\textbf{#1}}
\newcommand{\second}[1]{\underline{#1}}
\renewcommand{\arraystretch}{1.12}
\newcommand{\rot}[1]{\rotatebox{90}{#1}}

\renewcommand{\arraystretch}{1.12}

\newcommand{\QATableSetup}{%
  \footnotesize
  \setlength{\tabcolsep}{3.6pt}%
  \renewcommand{\arraystretch}{1.15}%
}

\DefineVerbatimEnvironment{MyVerbatim}{Verbatim}{
  breaklines=true,
  breakanywhere=true,
  fontsize=\footnotesize,
  frame=single,
  framesep=2mm
}

\newcommand{\DeltaShade}[1]{%
  \ifdim#1pt>30pt \cellcolor{green!30}\else
  \ifdim#1pt>20pt \cellcolor{green!24}\else
  \ifdim#1pt>10pt \cellcolor{green!18}\else
  \ifdim#1pt>2pt  \cellcolor{green!12}\else
  \ifdim#1pt>0pt  \cellcolor{green!8}\else
  \ifdim#1pt<-30pt \cellcolor{red!30}\else
  \ifdim#1pt<-20pt \cellcolor{red!24}\else
  \ifdim#1pt<-10pt \cellcolor{red!18}\else
  \ifdim#1pt<-2pt  \cellcolor{red!12}\else
  \ifdim#1pt<0pt   \cellcolor{red!8}\else
  \cellcolor{gray!10}%
  \fi\fi\fi\fi\fi\fi\fi\fi\fi\fi
}

\newcommand{\ScoreDelta}[2]{%
  \begingroup
  \DeltaShade{#2}%
  \shortstack{#1\\{\tiny\ifdim#2pt>0pt +#2\else #2\fi}}%
  \endgroup
}

\begin{document}

\title{DenTab: A Dataset for Table Recognition and Visual QA on Real-World Dental Estimates}
\author{Laziz Hamdi\inst{1,2}, Amine Tamasna\inst{2} \and Thierry Paquet\inst{1}}
\authorrunning{L. Hamdi}
\institute{LITIS, Normandy, France \and Malakoff Humanis, Paris, France}

\maketitle

\begin{abstract}
Tables condense key transactional and administrative information into compact layouts, but practical extraction requires more than text recognition: systems must also recover structure (rows, columns, merged cells, headers) and interpret roles such as line items, subtotals, and totals under common capture artifacts. Many existing resources for table structure recognition and TableVQA are built from clean digital-born sources or rendered tables, and therefore only partially reflect noisy administrative conditions.

We introduce DenTab, a dataset of 2{,}000 cropped table images from dental estimates with high-quality HTML annotations, enabling evaluation of table recognition (TR) and table visual question answering (TableVQA) on the same inputs. DenTab includes 2{,}208 questions across eleven categories spanning retrieval, aggregation, and logic/consistency checks. We benchmark 16 systems, including 14 vision--language models (VLMs) and two OCR baselines. Across models, strong structure recovery does not consistently translate into reliable performance on multi-step arithmetic and consistency questions, and these reasoning failures persist even when using ground-truth HTML table inputs.

To improve arithmetic reliability without training, we propose the Table Router Pipeline, which routes arithmetic questions to deterministic execution. The pipeline combines (i) a VLM that produces a baseline answer, a structured table representation, and a constrained table program with (ii) a rule-based executor that performs exact computation over the parsed table. The source code and dataset will be made publicly available at \url{https://github.com/hamdilaziz/DenTab}.
\end{abstract}

\section{Introduction}
Tables are a common format for transactional facts such as items, quantities, and prices. Extracting them from document images goes beyond OCR: a system must recover the 2D layout (rows, columns, merged cells, headers) and interpret functional roles (line items, subtotals, totals) while handling real capture artifacts such as blur, skew, occlusion, compression, and uneven illumination.

Table structure recognition (TSR) has progressed rapidly thanks to large benchmarks, including PubTabNet~\cite{pubtabnet}, FinTabNet~\cite{fintabnet}, TableBank~\cite{tablebank}, SciTSR~\cite{chi2019complicated}, and PubTables-1M~\cite{tatr}. However, these datasets are largely derived from digital-born documents or clean renderings and therefore underrepresent the noise and layout variability encountered in administrative scans and smartphone captures. In parallel, table question answering has been extensively studied, but many benchmarks assume a machine-readable table as input. In TableVQA, the input is an image, yet many recent resources rely on rendered or screenshot-style tables (e.g., TableVQA-Bench~\cite{kim2024tablevqa}) or synthetic renderings at scale (e.g., MMTab used by Table-LLaVA~\cite{zheng2024multimodal}), which still differ from real-world administrative imagery.

In this work, we study \emph{table recognition} (TR) and \emph{table visual question answering} (TableVQA) from a \emph{cropped table image}. We focus on questions that require correct structure grounding: from direct retrieval (lookup by header/key, ordinal row queries) to multi-step arithmetic (sums, conditional sums) and consistency checks that compare independently derived quantities.

To reflect practical deployment conditions, we introduce \textbf{DenTab}, a dataset of 2{,}000 real-world table images from dental estimates with high-quality HTML annotations capturing both structure and cell content, including frequent spanning cells. On top of these same inputs, we derive a TableVQA benchmark with 2{,}208 questions across eleven categories spanning retrieval, aggregation, comparison, and logic/consistency checks. We benchmark 16 systems (14 vision--language models and two OCR baselines) and observe a recurring gap: strong TSR does not reliably translate into correct answers for multi-step arithmetic and consistency questions. Importantly, these reasoning errors persist even when the image is replaced by ground-truth HTML, which isolates reasoning from perception.

To improve arithmetic reliability without training, we propose the \emph{Table Router Pipeline}. The pipeline routes arithmetic/consistency questions to deterministic execution: a VLM produces (i) a baseline answer, (ii) a table serialization, and (iii) a constrained table program, which is executed by a rule-based interpreter to obtain an exact result. We keep the baseline answer when execution is invalid or ungrounded, making the approach robust to imperfect structure predictions.

\noindent\textbf{Contributions.}
\begin{itemize}
    \item We introduce DenTab, a real-world dental-estimate dataset of 2{,}000 table crops with high-quality HTML structure+content annotations.
    \item We derive a TableVQA benchmark of 2{,}208 questions in eleven categories covering retrieval, aggregation, comparison, and consistency checks.
    \item We provide a unified zero-shot evaluation of recent VLMs and OCR systems, including robustness to degradations and an oracle-HTML setting that separates perception from reasoning.
    \item We propose and evaluate a training-free Table Router Pipeline that improves arithmetic/consistency questions by routing them to deterministic execution.
\end{itemize}

\begin{figure*}[t]
\centering
\includegraphics[width=0.8\linewidth]{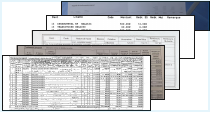}
\caption{Samples from DenTab showing diversity in table structures and real acquisition artifacts.}
\label{fig:devis_samples}
\end{figure*}

\section{Related Works}
\subsection{Datasets for table recognition and TableVQA}
\subsubsection{Table detection and table structure recognition}
Early benchmarks such as the ICDAR table competitions (cTDaR~\cite{gao2019icdar}) established standard evaluations for table detection and structure recognition, including recognition-only settings where the table crop is provided and only structure must be recovered. Large TSR datasets include TableBank~\cite{tablebank}, PubTabNet~\cite{pubtabnet}, SciTSR~\cite{chi2019complicated}, FinTabNet~\cite{fintabnet}, and PubTables-1M~\cite{tatr}. While influential, most are based on digital-born sources and therefore contain limited examples of real capture artifacts (blur, skew, compression) and administrative variability (irregular spacing, partial ruling lines, heterogeneous fonts), which are common in deployed document processing.

\subsubsection{Table QA/VQA}
Several widely used benchmarks evaluate reasoning over \emph{textual} tables, such as WikiTableQuestions~\cite{pasupat2015wikitq}, HiTab~\cite{cheng2022hitab}, TabFact~\cite{tabfact}, and TAT-QA~\cite{zhu2021tatqa}, and thus do not capture errors introduced by OCR and structure extraction. Document-level VQA datasets such as DocVQA~\cite{mathew2021docvqa} consider broader document understanding but are not designed to diagnose table structure recovery and table-specific reasoning. More table-focused visual benchmarks include TableVQA-Bench~\cite{kim2024tablevqa} and WikiDT~\cite{shi2024wikidt}, while multimodal table corpora such as MMTab (used by Table-LLaVA~\cite{zheng2024multimodal}) scale largely via rendering tables into images. Overall, existing resources often (i) underrepresent real capture conditions, and (ii) do not provide both HTML structure+content annotations and complex, use-case-driven questions on the same real table images, which makes it harder to analyze how structure errors propagate to reasoning failures.

\subsection{Table understanding methods}
Table understanding from document images is typically addressed by extracting structure and content and then answering downstream queries. Classical TSR relied on heuristics using ruling lines and whitespace. Learning-based approaches detect or segment table components and consolidate them into a grid, as in DeepDeSRT~\cite{deepdesrt}, TableNet~\cite{tablenet}, and CascadeTabNet~\cite{cascadetabnet}. End-to-end parsers also exist: PubTabNet popularized image-to-markup generation~\cite{pubtabnet}, GOT OCR~2.0~\cite{GOT2.0} integrates TSR with text extraction, and OCR-free document parsers such as Donut~\cite{donut} generate structured outputs directly.

More recently, vision--language models (VLMs) provide a unified interface by generating structured outputs or answers directly from images. General-purpose VLMs (e.g., Qwen-VL and successors~\cite{qwen_vl,bai2025qwen2,yang2025qwen3}, InternVL~\cite{internvl}) and instruction-tuned systems such as LLaVA~\cite{llava} can be prompted for TR and TableVQA, and large-context models (e.g., Gemma~3~\cite{team2025gemma}) are useful when serializing tables. Document-oriented OCR models and engines such as olmOCR~\cite{poznanski2025olmocr}, Nanonets-OCR~\cite{nanonetsocr,nanonetsocr2}, and Dolphin~\cite{feng2025dolphin} also support table extraction and question answering. Despite these advances, models can still make structural mistakes (notably spanning cells and header alignment) and may produce unreliable arithmetic even when the relevant values are correctly read.

\subsection{Tool-augmented methods for complex TableVQA}
Prompting methods such as Chain-of-Thought~\cite{wei2022chain} and self-consistency~\cite{ahmed2023better} can improve multi-step reasoning, but arithmetic is still generated by the model and remains error-prone. A complementary direction delegates computation to deterministic execution by having the model produce executable code or programs, such as PAL~\cite{gao2023pal} and Program-of-Thoughts~\cite{chen2022program}. In multimodal settings, visual-programming systems such as VisProg~\cite{gupta2023visual} and ViperGPT~\cite{suris2023vipergpt} generate programs that call vision modules and execute them. ReAct~\cite{yao2022react} and MM-ReAct~\cite{yang2023mm} interleave reasoning with tool calls, while Toolformer~\cite{schick2023toolformer} studies learning when to use tools. DePlot~\cite{liu2023deplot} converts plots into table-like text representations to support downstream reasoning. Our work follows the tool-augmentation principle for real table images by routing arithmetic/consistency questions to deterministic execution of constrained table programs grounded in an extracted table representation.

\section{DenTab Dataset}
DenTab is a real-world dataset of 2{,}000 tables extracted from french administrative dental estimates, designed for table recognition (TR), table structure recognition (TSR), and table visual question answering (TableVQA) on the \emph{same} table images. The data are split into 1{,}200/400/400 for training/validation/test.

Each table crop is annotated with an HTML representation capturing cell content, row/column structure, and spanning cells. An in-house web annotation tool supporting row/column editing and merged-cell operations has been used by four expert annotators to produce the ground truth, following detailed guidelines (Appendix). DenTab targets real administrative captures where blur, compression, and layout variability are common, in contrast to existing large-scale datasets.

\begin{figure}[t]
\centering
\includegraphics[width=0.98\linewidth]{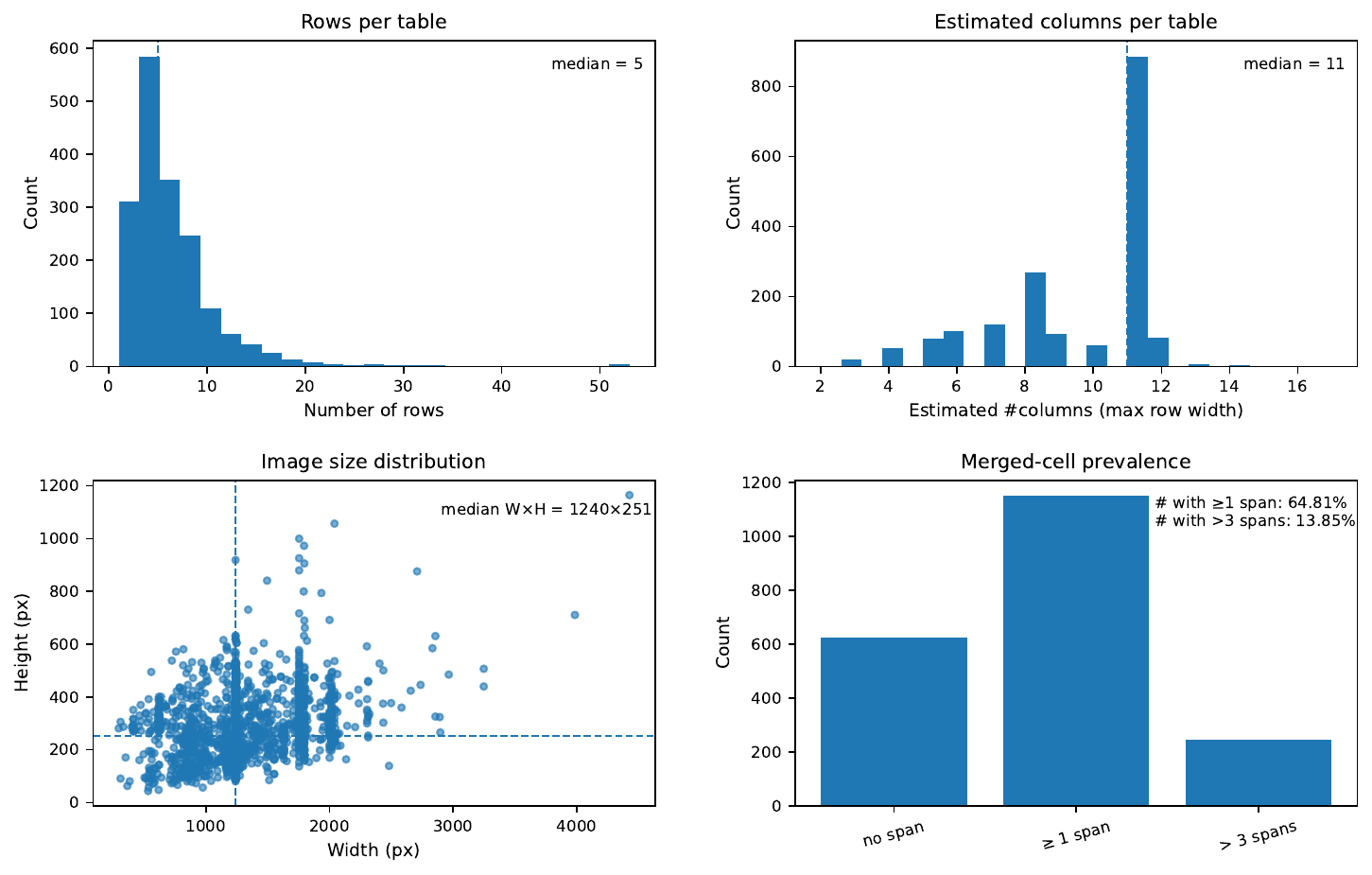}
\caption{Overview statistics of DenTab: distributions of rows per table and columns per table, image size (width vs.\ height), and prevalence of spanning (merged) cells.}
\label{fig:dataset_overview}
\end{figure}

Figure~\ref{fig:dataset_overview} summarizes the dataset properties. The median crop size is 1240$\times$251 px (width$\times$height), with a median of 5 rows and 11 columns. Merged cells are frequent: 64.81\% of tables contain at least one spanning cell and 13.85\% contain more than three, making accurate span reconstruction a key challenge.

DenTab also provides TableVQA annotations, where the input is a cropped table image and a natural-language question, and the output is a short textual or numerical answer. Unlike table QA datasets that operate on \emph{textual} tables (e.g., WikiTableQuestions, TabFact, TAT-QA, HiTab)~\cite{cheng2022hitab,tabfact,zhu2021tatqa,cheng2022hitab}, DenTab evaluates QA under visual noise and noisy structure. Unlike document-general VQA (DocVQA)~\cite{mathew2021docvqa} datasets, DenTab focuses on table-centric queries that depend on correct row/column alignment and cell spanning. Compared to recent TableVQA resources (e.g., TableVQA-Bench) and large multimodal table corpora that rely heavily on rendered/screenshot-style tables (e.g., MMTab/Table-LLaVA), DenTab uses real administrative captures with realistic noise.

DenTab comprises eleven visual questions categories as detailed in Table~\ref{tab:qa_categories},  spanning: (i) \emph{Retrieval} (lookup by header/key, list retrieval, ordinal row queries, and interpreting missing cells as N/A), (ii) \emph{Aggregation} (reading explicit totals versus computing sums, including conditional sums), and (iii) \emph{Logic and consistency} (argmax queries, counting, and consistency checks that compare independently derived quantities). See appendix for representative examples. This taxonomy separates direct ``read-a-cell'' questions from structure-grounded operations.
\begin{table}[t]
\caption{DenTab TableVQA categories ($X,Y$ are headers). Categories marked with \textbf{\(\star\)} are treated as \emph{complex arithmetic} and are targeted by the program-execution branch of our router pipeline.}
\label{tab:qa_categories}
\centering
\small
\setlength{\tabcolsep}{3pt}
\begin{tabular}{p{4.8cm}p{6.9cm}}
\toprule
\textbf{Category} & \textbf{Description} \\
\midrule
\multicolumn{2}{l}{\emph{Retrieval}}\\
\texttt{lookup\_by\_header} & Value in column $Y$ for the row where column $X$ equals a key (unique).\\
\texttt{lookup\_list\_by\_header} & Same, but return all matching rows.\\
\texttt{kth\_row\_value} & Value in row $k$ (optionally data-only).\\
\texttt{na\_from\_empty} & Treat empty cell as N/A when required.\\
\midrule
\multicolumn{2}{l}{\emph{Aggregation}}\\
\texttt{total\_row\_value} & Read value from a total/subtotal row.\\
\texttt{aggregation\_sum} \textbf{\(\star\)} & Sum column values over data rows.\\
\texttt{aggregation\_sum\_conditional} \textbf{\(\star\)} & Sum over rows matching a condition.\\
\midrule
\multicolumn{2}{l}{\emph{Logic \& consistency}}\\
\texttt{comparison\_argmax} & Return key(s) for maximum in a column (ties allowed).\\
\texttt{comparison\_argmax\_rows} & Return row index/indices of the maximum.\\
\texttt{count\_equals} \textbf{\(\star\)} & Count occurrences of a value (or rows in context).\\
\texttt{consistency\_diff\_total} \textbf{\(\star\)} & Difference between two derived quantities (e.g., sum$A$--sum$B$).\\
\bottomrule
\end{tabular}
\end{table}

\section{Table Router Pipeline for Arithmetic TableVQA}
Vision--language models (VLMs) can answer many table questions directly from images, but they often struggle when the answer requires \emph{exact} computation (e.g., sums, differences, or counts). While prompting techniques such as chain-of-thought or self-consistency can improve reasoning, the final arithmetic is still produced by the model and can therefore be unreliable. To address this, we adopt a program-assisted approach: the VLM generates an explicit program, which is then executed by a deterministic interpreter to produce the numerical result.

\begin{figure}[t]
\centering
\includegraphics[width=0.7\linewidth]{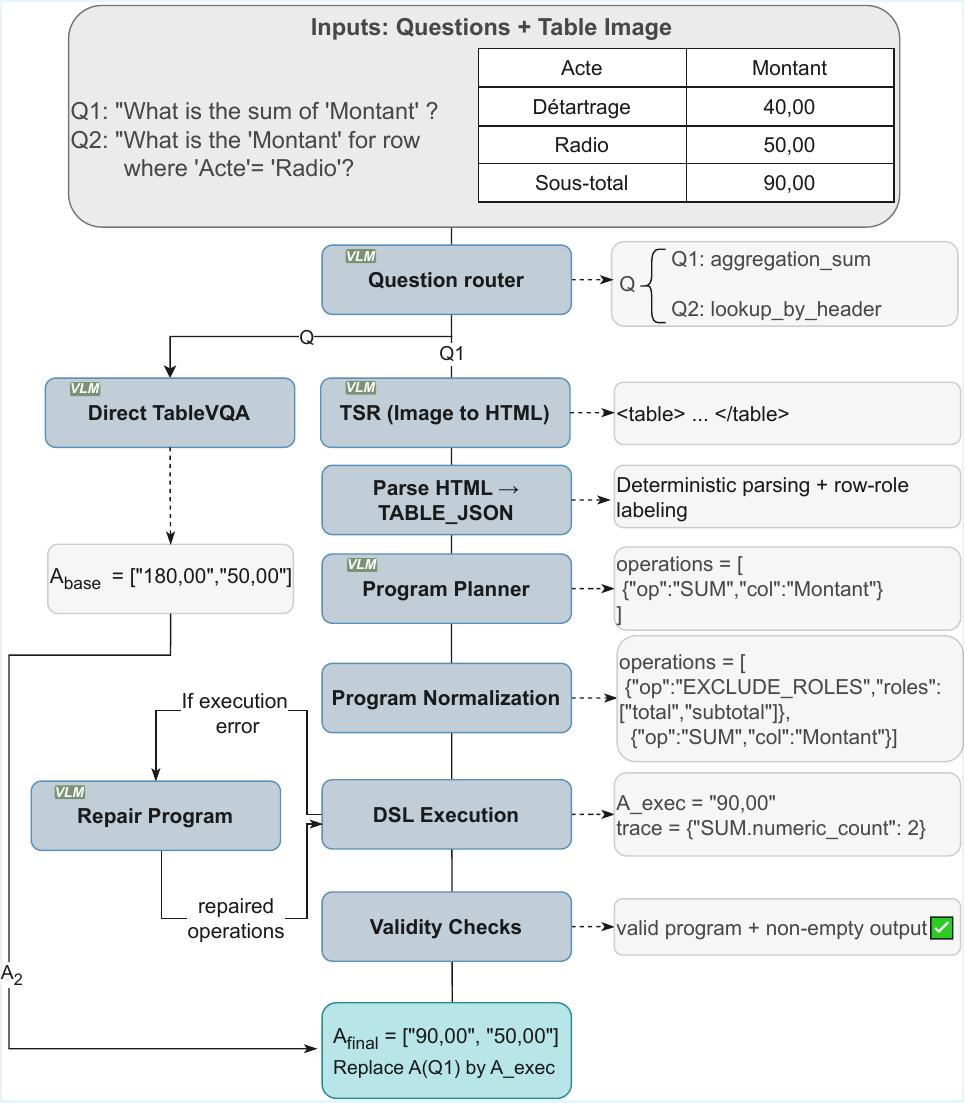}
\caption{\textbf{Table Router Pipeline for arithmetic TableVQA.} All questions are answered directly from the image to produce baseline outputs $A_{\text{base}}$. In parallel, the router selects arithmetic questions (e.g., $Q_1$) for program-assisted execution: TSR converts the image to HTML, which is parsed into \texttt{TABLE\_JSON}; a VLM generates a DSL program that is normalized (e.g., excluding total/subtotal rows) and executed deterministically. If validity checks succeed, the executed answer $a_{\text{exec}}$ replaces the corresponding baseline entry (here $180{,}00 \rightarrow 90{,}00$), yielding $A_{\text{final}}$; otherwise the baseline answer is kept. A repair step is optionally invoked on execution errors.}
\label{fig:table_router}
\end{figure}

Figure~\ref{fig:table_router} illustrates the end-to-end pipeline on a simple example. Given a table image and a set of questions $Q$ (e.g., $Q_1$ asks for an aggregation, while $Q_2$ is a simple lookup), we first apply a \emph{question router} that assigns each question to one category from our taxonomy (Table~\ref{tab:qa_categories}). After that all questions are sent to a \emph{Direct TableVQA} branch, which produces baseline answers $A_{\text{base}}$ by prompting the VLM to answer directly from the image.

For questions routed to arithmetic categories (e.g., $Q_1$ in Figure~\ref{fig:table_router}), we additionally invoke the program-execution branch. We (i) reconstruct the table as HTML using a TSR model, (ii) deterministically parse the HTML into a normalized table representation (\texttt{TABLE\_JSON}), and (iii) prompt a \emph{program planner} to output a short program in a domain-specific language (DSL). The DSL is intentionally restricted to table operations such as row/column selection, filtering, and terminal operators (e.g., \texttt{SUM}, \texttt{COUNT}, \texttt{DIFF}). Before execution, we apply rule-based program normalization (e.g., excluding \texttt{total}/\texttt{subtotal} rows for non-total aggregations). The resulting program is executed by a deterministic interpreter that supports financial number formats and exact decimal arithmetic, producing an execution answer $A_{\text{exec}}$ along with an execution trace.

We use validity checks to decide when to trust program execution. Concretely, we override the corresponding entry in $A_{\text{base}}$ with $a_{\text{exec}}$ only if execution succeeds and yields a non-empty grounded output. In Figure~\ref{fig:table_router}, the direct answer for $Q_1$ is corrected from $180{,}00$ to $90{,}00$ after normalization removes the subtotal row, resulting in the final answer set $A_{\text{final}}$. If execution fails (e.g., malformed DSL or missing column references), we optionally trigger a \emph{repair} step by re-prompting the VLM with the error message and execution trace to generate a corrected program.


\section{Experiments and Results}
For TSR, we report three complementary families of metrics. TEDS measures the tree-edit similarity between predicted and reference HTML by comparing their DOM trees with text content, and S-TEDS is the structure only variant~\cite{pubtabnet}. To capture local structural consistency, we also compute adjacency precision/recall/F1 from the cell neighborhood graph: each cell induces directed adjacency relations (right/left and up/down neighbors) in the reconstructed grid, and we compare the predicted and ground-truth adjacency edge sets. Finally, we report GRITS$_{\text{top}}$~\cite{smock2023grits}, a topology-focused score that evaluates grid structure recovery independently of textual content. 

We benchmark 16 systems: 14 vision--language models (VLMs) and 2 OCR baselines. The evaluated VLMs include general-purpose multimodal models from the Qwen family (Qwen2.5-VL and Qwen3-VL), Gemma-3, DeepSeek-VL2, Idef3, and Ministral-3, as well as document/OCR-focused models (olmOCR-2 and Dolphin-v2). As OCR baselines, we report Nanonets-OCR-s and Nanonets-OCR2-3B. The models span multiple parameter scales, ranging from small models ($\leq$5B) to mid-size models (7B--15B) and large models ($\geq$20B). Note that Qwen3-VL-30B-A3B is a mixture-of-experts checkpoint; under Qwen's A3B naming convention, only a subset of parameters is activated per token. For reproducibility, the Appendix reports the exact checkpoint IDs (including suffixes such as \texttt{-1025} or \texttt{-2512}), as well as prompts and hyperparameters used in all experiments see appendix "Prompts, Hyperparameters, and Model Versions". All the following results are obtained on the test set images.
\subsection{Table Structure Recognition}

\begin{table*}[t]
\centering
\small
\setlength{\tabcolsep}{3.8pt}
\caption{TSR results on the dental estimates dataset (\%, higher is better). Models are grouped by parameter scale and sorted by S-TEDS within each group. Best and second-best values per column are highlighted.}
\label{tab:steds_results}
\begingroup
\fontsize{8}{10}\selectfont
\begin{tabular}{l|c|cccccc}
\toprule
Model & Size & S-TEDS & TEDS & Prec. & Rec. & F1 & GRITS$_{\text{top}}$ \\
\midrule

\multicolumn{8}{l}{\textit{Small models ($\leq$ 5B)}}\\
\midrule
Nanonets-OCR2~\cite{nanonetsocr2}      & 3B   & \second{90.18} & \second{85.55} & \second{74.52} & \second{74.23} & \second{74.27} & \second{78.77} \\
Qwen3-VL~\cite{yang2025qwen3}          & 4B   & 87.39 & 82.47 & 68.03 & 67.69 & 67.67 & 73.41 \\
Nanonets-OCR~\cite{nanonetsocr}        & 3B   & 85.55 & 80.72 & 71.74 & 71.69 & 71.45 & 69.05 \\
DeepSeek-VL2~\cite{wu2024deepseek}     & 4B   & 79.16 & 66.72 & 49.63 & 48.82 & 48.52 & 52.83 \\
Gemma-3~\cite{team2025gemma}           & 4B   & 72.07 & 56.91 & 36.11 & 31.48 & 32.97 & 44.92 \\
Dolphin-v2~\cite{feng2025dolphin}      & 4B   & 65.59 & 57.49 & 48.40 & 46.82 & 46.56 & 53.94 \\

\midrule
\multicolumn{8}{l}{\textit{Mid-size models (7B--15B)}}\\
\midrule
Qwen3-VL~\cite{yang2025qwen3}          & 8B   & 89.50 & 85.05 & 73.19 & 73.25 & 73.14 & 69.89 \\
OlmOCR-2~\cite{poznanski2025olmocr}    & 7B   & 84.36 & 79.22 & 72.71 & 71.94 & 72.16 & 70.64 \\
Ministral-3~\cite{mistral2024mistral3} & 8B  & 82.68 & 75.28 & 64.10 & 62.10 & 62.90 & 57.19 \\
Ministral-3~\cite{mistral2024mistral3} & 14B & 81.59 & 74.50 & 63.56 & 62.43 & 62.86 & 56.47 \\
Gemma-3~\cite{team2025gemma}           & 12B  & 80.78 & 72.47 & 49.12 & 46.97 & 47.88 & 60.65 \\
Qwen2.5-VL~\cite{bai2025qwen2}         & 7B   & 79.79 & 74.29 & 67.66 & 66.76 & 67.04 & 65.59 \\
Idef3~\cite{laurencon2024building}     & 8B   & 68.63 & 47.95 & 20.24 & 19.90 & 19.37 & 42.93 \\

\midrule
\multicolumn{8}{l}{\textit{Large models ($\geq$ 20B)}}\\
\midrule
Qwen3-VL~\cite{yang2025qwen3}          & 32B  & \best{92.70} & \best{89.70} & \best{75.28} & \best{75.56} & \best{75.30} & \best{86.10} \\
Qwen3-VL-A3B~\cite{yang2025qwen3}      & 30B  & 86.64 & 82.03 & 71.60 & 72.85 & 72.08 & 69.26 \\
Gemma-3~\cite{team2025gemma}           & 27B  & 82.40 & 75.58 & 58.57 & 57.27 & 57.75 & 65.82 \\
\bottomrule
\end{tabular}
\endgroup
\end{table*}

Table~\ref{tab:steds_results} shows substantial variation across model families. The strongest TSR performance is obtained by Qwen3-VL-32B, which achieves the best score on every metric (92.70 S-TEDS, 89.70 TEDS, 75.30 F1, and 86.10 GRITS$_{\text{top}}$), indicating highly accurate span recovery and consistent grid topology at scale.

Nanonets-OCR2-3B remains a close second overall (90.18 S-TEDS and 85.55 TEDS, with 74.27 F1 and 78.77 GRITS$_{\text{top}}$), and is the strongest model in the small-model regime. This aligns with its objective: Nanonets-OCR2 is optimized for converting document images into structured outputs under document-centric supervision that closely matches TSR. Within the Qwen family, performance increases with scale: Qwen3-VL-8B is strong among mid-size models (89.50 S-TEDS and 85.05 TEDS), while Qwen3-VL-32B provides a clear additional gain. At the same time, scale alone does not guarantee better TSR: the MoE checkpoint Qwen3-VL-A3B-30B reaches 86.64 S-TEDS and 72.08 F1, well below the dense 32B model and also behind the best document-specialized system. Overall, both model capacity and training objective matter: larger dense VLMs can lead on TSR, but document-focused training remains highly competitive, especially at smaller scales. (See predictions samples in the appendix "Table Recognition samples").

\subsection{Table Question Answering Results}
For Table QA, we report the subset of models that reliably support direct-answer inference under our evaluation constraints; TSR results include the full set of 16 systems. In our analysis, we define 11 question categories, but we omit the \texttt{na from empty} category from the main tables to keep them readable; results for this category are reported in the supplementary material.

\subsubsection{Zero-Shot Table QA from Images}
Given a table image and a question, each model outputs a final answer; we compute exact-match accuracy per category. For readability and consistency with TSR, all accuracies are reported in percent (\%).

\begin{table*}[t]
\centering
\begingroup
\QATableSetup
\caption{Exact-match accuracy on TableVQA (\%). Best and second-best values per column are highlighted.\\
\protect\\[-0.3em]{\footnotesize \textbf{Abbrev.} Q25-7B=Qwen2.5-VL-7B; Q3-4B/Q3-8B/Q3-30B/Q3-32B=Qwen3-VL-4B/8B/30B-A3B/32B; Olm=olmOCR-2-7B; NN-s=Nanonets-OCR-s; NN2-3B=Nanonets-OCR2-3B; Idef3=Idefics3-8B; G3-4B/G3-12B/G3-27B=Gemma-3-4B/12B/27B; M3-8B=Ministral-3-8B.
Sum=aggregation sum; Sum-C=aggregation sum conditional; Total=total row value; ArgMax=comparison argmax; ArgMax-R=comparison argmax rows; Diff=consistency diff total; Eq=count equals; Lookup=lookup by header; Lookup-L=lookup list by header; Kth=kth row value. In Table~\ref{tab:vqa_results}, parentheses denote \#questions.}}
\label{tab:vqa_results}

\begin{tabular}{l|ccc|cccc|ccc|c}
\toprule
 & \multicolumn{3}{c}{\textbf{Aggregation}} & \multicolumn{4}{c}{\textbf{Logic}} & \multicolumn{3}{c}{\textbf{Retrieval}} & \textbf{Overall} \\
Model
& \rot{\shortstack{Sum\\{\scriptsize (205)}}}
& \rot{\shortstack{Sum-C\\{\scriptsize (185)}}}
& \rot{\shortstack{Total\\{\scriptsize (169)}}}
& \rot{\shortstack{ArgMax\\{\scriptsize (239)}}}
& \rot{\shortstack{ArgMax-R\\{\scriptsize (75)}}}
& \rot{\shortstack{Diff\\{\scriptsize (198)}}}
& \rot{\shortstack{Eq\\{\scriptsize (255)}}}
& \rot{\shortstack{Lookup\\{\scriptsize (256)}}}
& \rot{\shortstack{Lookup-L\\{\scriptsize (191)}}}
& \rot{\shortstack{Kth\\{\scriptsize (256)}}}
& \rot{\shortstack{Overall\\{\scriptsize (2208)}}} \\
\midrule

Q25-7B
& 51.7 & 42.6 & 81.2 & 48.7 & 21.3 & 11.7 & 85.7 & 70.6 & 45.2 & 78.2 & 56.0 \\
Q3-4B
& 57.6 & 40.4 & 90.9 & 45.3 & 18.7 & 11.7 & 88.4 & 76.2 & 34.6 & 81.7 & 55.8 \\
Q3-8B
& 62.1 & 45.9 & 89.7 & 55.5 & 29.3 & 16.3 & 88.4 & \second{77.4} & \second{60.1} & 86.9 & 61.7 \\
Q3-30B
& \second{65.0} & \second{67.2} & \second{92.7} & \best{62.7} & 44.0 & \second{25.0} & 90.8 & 71.4 & 50.5 & \best{90.9} & \second{66.2} \\
Q3-32B
& \best{68.5} & \best{67.8} & \best{93.3} & \second{61.9} & \best{53.3} & \best{48.5} & \best{93.6} & \best{82.9} & \best{64.9} & \second{89.3} & \best{73.2} \\
Olm
& 56.7 & 32.2 & 89.7 & 41.5 & 21.3 & 9.2 & 85.3 & 70.2 & 47.3 & 78.2 & 54.5 \\
NN-s
& 14.3 & 7.7 & 37.0 & 30.5 & 10.7 & 5.6 & 37.1 & 53.6 & 14.9 & 58.7 & 29.9 \\
NN2-3B
& 27.6 & 16.4 & 27.9 & 19.1 & 4.0 & 7.7 & 27.1 & 52.4 & 11.7 & 46.8 & 27.5 \\
Idef3
& 25.1 & 8.2 & 38.2 & 25.8 & 6.7 & 7.1 & 50.6 & 45.2 & 6.9 & 53.6 & 30.1 \\
G3-4B
& 34.5 & 16.9 & 59.4 & 34.3 & 22.7 & 6.1 & 55.4 & 52.0 & 5.9 & 38.9 & 32.3 \\
G3-12B
& 53.2 & 38.3 & 78.8 & 52.1 & 26.7 & 9.2 & 89.6 & 65.9 & 33.0 & 73.4 & 52.2 \\
G3-27B
& 58.1 & 48.1 & 82.4 & 59.3 & \second{48.0} & 13.3 & \second{92.0} & 66.7 & 37.2 & 76.2 & 56.6 \\
M3-8B
& 36.5 & 38.8 & 50.9 & 48.7 & 33.3 & 11.7 & 66.9 & 54.8 & 35.1 & 59.5 & 44.9 \\

\bottomrule
\end{tabular}
\endgroup
\end{table*}

Table~\ref{tab:vqa_results} shows that Q3-32B achieves the best overall accuracy (73.2\%), followed by Q3-30B (66.2\%). The strongest gains appear in the hardest logic categories: \texttt{consistency diff total} reaches 48.5\% (vs.\ 25.0\% for the next best, Q3-30B), and \texttt{comparison argmax rows} peaks at 53.3\%. Retrieval remains comparatively robust (e.g., \texttt{lookup list by header} peaks at 64.9\% and \texttt{kth row value} reaches 90.9\%), while multi-step global reasoning still accounts for most remaining errors. OCR-centric systems (Olm and Nanonets) trail the strongest VLMs overall, suggesting that end-to-end multimodal reasoning provides a clearer advantage than OCR alone on this benchmark, see appendix "Direct TableVQA" for prediction samples.

\subsubsection{Oracle HTML Table QA (Perception-Free)}
\label{subsub:html+tqa}
To isolate reasoning errors from perception errors, we replace the image with the ground-truth HTML table serialization. Table~\ref{tab:vqa_html_oracle} shows large gains across most categories, with the strongest improvements in retrieval. Several models approach saturation on \texttt{lookup by header}, peaking at 99.2\% (M3-14B), which indicates that a substantial fraction of errors in the image-based setting originates from imperfect table reading (OCR noise, missed spans, and header misalignment).

However, removing perception does not eliminate the difficulty of numerical reasoning. In particular, \texttt{consistency diff total} remains the lowest-performing category for most models, although Q3-32B improves markedly to 55.1\% (vs.\ 44.9\% for the next best, Q3-30B). Overall, Q3-32B achieves the strongest oracle accuracy (83.4\%), closely followed by Q3-30B (82.6\%) and M3-14B (82.0\%). Q3-30B performs best on several arithmetic and comparison categories (e.g., \texttt{aggregation sum}, \texttt{total row value}, \texttt{comparison argmax}, \texttt{comparison argmax rows}, and \texttt{kth row value}), while M3-14B is strongest on retrieval (\texttt{lookup by header}=99.2\%, \texttt{lookup list by header}=88.3\%).

\begin{table*}[t]
\centering
\begingroup
\QATableSetup
\caption{Oracle-HTML Table QA results (\% exact match). Best and second-best values per column are highlighted.\\
\protect\\[-0.3em]{\footnotesize \textbf{Abbrev.} M3-14B=Ministral-3-14B.}}
\label{tab:vqa_html_oracle}

\begin{tabular}{l|ccc|cccc|ccc|c}
\toprule
 & \multicolumn{3}{c}{\textbf{Aggregation}} & \multicolumn{4}{c}{\textbf{Logic}} & \multicolumn{3}{c}{\textbf{Retrieval}} & \textbf{Overall} \\
Model & \rot{Sum} & \rot{Sum-C} & \rot{Total} & \rot{ArgMax} & \rot{ArgMax-R} & \rot{Diff} & \rot{Eq} & \rot{Lookup} & \rot{Lookup-L} & \rot{Kth} & \rot{Overall} \\
\midrule
Q25-7B  & 55.2 & 36.6 & 83.6 & 43.6 & 18.7 & 9.7  & 88.4 & 92.9 & 47.3 & 86.9 & 63.2 \\
Q3-4B   & 64.5 & 56.8 & 91.5 & 56.8 & 34.7 & 17.3 & 90.0 & 97.2 & 68.1 & \second{96.0} & 72.2 \\
Q3-8B   & 66.5 & 68.3 & 87.9 & 63.1 & 40.0 & 18.9 & 90.0 & 97.6 & \second{80.3} & 93.3 & 74.5 \\
Q3-30B  & \best{78.8} & \second{83.6} & \best{96.4} & \best{70.3} & \best{69.3} & \second{44.9} & \best{95.6} & 94.0 & \second{80.3} & \best{96.4} & \second{82.6} \\
Q3-32B  & \second{76.8} & \best{84.7} & \second{94.5} & 68.2 & 64.0 & \best{55.1} & \second{94.4} & 97.2 & 79.3 & 93.7 & \best{83.4} \\
Olm     & 52.2 & 35.5 & 87.3 & 47.0 & 24.0 & 15.3 & 85.3 & 87.3 & 58.0 & 82.9 & 62.7 \\
NN-s    & 28.6 & 12.0 & 42.4 & 41.5 & 20.0 & 3.1  & 36.3 & 68.7 & 16.0 & 60.3 & 38.4 \\
NN2-3B  & 25.1 & 15.8 & 26.1 & 25.8 & 5.3  & 5.1  & 40.2 & 57.5 & 8.0  & 53.6 & 32.3 \\
Idef3   & 37.9 & 21.9 & 47.3 & 48.7 & 21.3 & 5.6  & 78.9 & 87.3 & 19.7 & 83.3 & 52.7 \\
G3-4B   & 47.8 & 22.4 & 35.8 & 38.1 & 16.0 & 6.1  & 59.0 & 72.6 & 13.3 & 55.6 & 42.0 \\
G3-12B  & 50.7 & 54.1 & 84.8 & 57.2 & 25.3 & 11.7 & 93.6 & 98.4 & 59.6 & 92.1 & 69.0 \\
G3-27B  & 63.5 & 70.5 & 88.5 & 67.4 & 57.3 & 22.4 & 92.0 & 97.2 & 61.7 & 95.2 & 75.4 \\
M3-8B   & 61.1 & 69.4 & 73.9 & \second{69.5} & 58.7 & 32.1 & 88.8 & \second{98.8} & 75.0 & 89.3 & 75.2 \\
M3-14B  & 70.9 & \second{83.6} & 87.3 & 67.8 & \second{66.7} & 42.9 & 93.6 & \best{99.2} & \best{88.3} & 94.8 & 82.0 \\
\bottomrule
\end{tabular}
\endgroup
\end{table*}

In this oracle setting, retrieval categories approach saturation for the strongest models (e.g., \texttt{lookup by header} reaches 99.2\% with M3-14B). Nevertheless, numerical reasoning remains substantially harder: the best accuracy for \texttt{consistency diff total} is 55.1\% (Q3-32B), indicating that multi-step arithmetic and constraint adherence remain a major source of errors even when perception is removed. Overall, Q3-32B achieves the best oracle-HTML accuracy (83.4\%), closely followed by Q3-30B (82.6\%).

\subsection{Table Router Pipeline Evaluation}
Due to space constraints, we report a subset of models here; full results are in the supplementary material.

\subsubsection{Table Router Pipeline from Images}
\begin{table*}[t]
\centering
\begingroup
\QATableSetup
\caption{Exact-match accuracy on TableVQA (\%). Each cell shows score and $\Delta$ (percentage-point change) vs.\ direct QA in Table~\ref{tab:vqa_results}; green indicates improvements and red indicates drops. Overall is the weighted average over the 10 shown categories (Sum..Kth) using the per-category counts from Table~\ref{tab:vqa_results}; Overall $\Delta$ is computed against the same 10-category weighted average from Table~\ref{tab:vqa_results} (excluding \texttt{na from empty}).}
\label{tab:vqa_results_delta_new}

\begin{tabular}{l|ccc|cccc|ccc|c}
\toprule
 & \multicolumn{3}{c}{\textbf{Aggregation}} & \multicolumn{4}{c}{\textbf{Logic}} & \multicolumn{3}{c}{\textbf{Retrieval}} & \textbf{Overall} \\
Model
& \rot{Sum} & \rot{Sum-C} & \rot{Total}
& \rot{ArgMax} & \rot{ArgMax-R} & \rot{Diff} & \rot{Eq}
& \rot{Lookup} & \rot{Lookup-L} & \rot{Kth}
& \rot{Overall} \\
\midrule

Q25-7B
& \ScoreDelta{64.4}{35.0}
& \ScoreDelta{42.3}{0.2}
& \ScoreDelta{79.9}{6.3}
& \ScoreDelta{48.3}{2.0}
& \ScoreDelta{20.9}{-1.0}
& \ScoreDelta{12.6}{7.0}
& \ScoreDelta{85.8}{-0.2}
& \ScoreDelta{69.9}{-0.2}
& \ScoreDelta{44.6}{1.4}
& \ScoreDelta{78.1}{1.8}
& \ScoreDelta{58.4}{3.0} \\

Q3-4B
& \ScoreDelta{73.2}{24.3}
& \ScoreDelta{74.6}{29.7}
& \ScoreDelta{89.7}{-0.6}
& \ScoreDelta{52.3}{5.5}
& \ScoreDelta{26.7}{2.6}
& \ScoreDelta{38.9}{42.8}
& \ScoreDelta{88.6}{4.9}
& \ScoreDelta{82.0}{0.0}
& \ScoreDelta{50.3}{-1.1}
& \ScoreDelta{84.0}{-0.4}
& \ScoreDelta{69.4}{9.4} \\

Q3-8B
& \ScoreDelta{72.2}{24.2}
& \ScoreDelta{80.5}{30.1}
& \ScoreDelta{88.6}{1.4}
& \ScoreDelta{61.5}{4.3}
& \ScoreDelta{30.7}{-0.3}
& \ScoreDelta{38.9}{41.2}
& \ScoreDelta{88.3}{7.6}
& \ScoreDelta{\second{86.3}}{0.0}
& \ScoreDelta{\second{64.9}}{-1.3}
& \ScoreDelta{86.7}{0.8}
& \ScoreDelta{73.2}{10.2} \\

Q3-30B
& \ScoreDelta{\second{74.6}}{13.9}
& \ScoreDelta{\second{81.1}}{9.9}
& \ScoreDelta{\second{91.8}}{-0.7}
& \ScoreDelta{\second{64.4}}{8.4}
& \ScoreDelta{\best{65.3}}{1.4}
& \ScoreDelta{\second{46.0}}{17.2}
& \ScoreDelta{\second{90.8}}{1.3}
& \ScoreDelta{80.1}{2.1}
& \ScoreDelta{63.9}{-0.2}
& \ScoreDelta{\best{90.6}}{0.1}
& \ScoreDelta{\second{76.0}}{6.0} \\

Q3-32B
& \ScoreDelta{\best{82.9}}{19.1}
& \ScoreDelta{\best{87.6}}{11.0}
& \ScoreDelta{\best{92.5}}{1.9}
& \ScoreDelta{\best{66.1}}{3.8}
& \ScoreDelta{\second{54.7}}{1.3}
& \ScoreDelta{\best{50.5}}{6.5}
& \ScoreDelta{\best{93.6}}{3.2}
& \ScoreDelta{\best{91.0}}{2.0}
& \ScoreDelta{\best{68.6}}{0.3}
& \ScoreDelta{\second{89.8}}{1.2}
& \ScoreDelta{\best{79.8}}{5.7} \\

Olm
& \ScoreDelta{66.8}{38.0}
& \ScoreDelta{33.0}{5.6}
& \ScoreDelta{88.2}{0.0}
& \ScoreDelta{51.0}{9.5}
& \ScoreDelta{21.1}{-1.1}
& \ScoreDelta{16.2}{9.4}
& \ScoreDelta{85.2}{-0.1}
& \ScoreDelta{72.7}{-0.3}
& \ScoreDelta{46.5}{-1.2}
& \ScoreDelta{78.2}{1.9}
& \ScoreDelta{59.6}{5.2} \\

NN-s
& \ScoreDelta{43.4}{23.1}
& \ScoreDelta{20.5}{8.0}
& \ScoreDelta{35.9}{0.4}
& \ScoreDelta{34.7}{-1.0}
& \ScoreDelta{10.6}{-1.5}
& \ScoreDelta{13.6}{12.6}
& \ScoreDelta{37.0}{-0.2}
& \ScoreDelta{53.7}{-1.3}
& \ScoreDelta{14.2}{-0.7}
& \ScoreDelta{58.6}{-0.1}
& \ScoreDelta{35.2}{1.3} \\

NN2-3B
& \ScoreDelta{44.4}{35.4}
& \ScoreDelta{16.6}{3.7}
& \ScoreDelta{27.3}{-1.4}
& \ScoreDelta{18.6}{-0.9}
& \ScoreDelta{3.9}{-0.2}
& \ScoreDelta{16.7}{27.7}
& \ScoreDelta{26.5}{-2.3}
& \ScoreDelta{51.1}{-2.5}
& \ScoreDelta{11.0}{-0.6}
& \ScoreDelta{45.8}{-2.6}
& \ScoreDelta{28.8}{2.8} \\

Idef3
& \ScoreDelta{26.3}{53.3}
& \ScoreDelta{16.8}{51.6}
& \ScoreDelta{37.6}{-1.7}
& \ScoreDelta{28.5}{-0.2}
& \ScoreDelta{12.0}{-0.8}
& \ScoreDelta{8.6}{56.0}
& \ScoreDelta{51.3}{8.9}
& \ScoreDelta{44.7}{-0.5}
& \ScoreDelta{7.2}{0.1}
& \ScoreDelta{54.1}{1.9}
& \ScoreDelta{31.6}{16.1} \\

\bottomrule
\end{tabular}
\endgroup
\end{table*}
Table~\ref{tab:vqa_results_delta_new} shows that the router pipeline substantially boosts performance on aggregation and multi-step reasoning for most models. The largest reported gains concentrate in \texttt{Sum}, \texttt{Sum-C}, and especially \texttt{Diff} (e.g., \texttt{Diff} increases by +42.8 for Q3-4B and +41.2 for Q3-8B), while changes in retrieval (\texttt{Lookup}, \texttt{Lookup-L}, \texttt{Kth}) are comparatively smaller and often mixed in sign. In absolute terms, Q3-32B remains strongest overall (79.8\%), followed by Q3-30B (76.0\%). The router also improves the hardest consistency checks: \texttt{Diff} rises to 50.5\% for Q3-32B (46.0\% for Q3-30B), though it remains the lowest-scoring category even after routing. Overall, these results suggest that explicit routing/program planning disproportionately benefits arithmetic-heavy queries, while retrieval and simpler comparisons are already relatively strong and see smaller net changes.

\subsubsection{On HTML annotations}
\begin{table*}[t]
\centering
\begingroup
\QATableSetup
\caption{Table Router Pipeline on HTML annotations (\% exact match). Overall is a weighted average using the per-category question counts from Table~\ref{tab:vqa_results}. We also show $\Delta$ (percentage-point change) vs.\ Question Answering from the annotated HTML from Table~\ref{tab:vqa_html_oracle}; green indicates improvements, red indicates drops.}
\label{tab:hvs_html_rooter}
\begin{tabular}{l|ccc|cccc|ccc|c}
\toprule
 & \multicolumn{3}{c}{Aggregation} & \multicolumn{4}{c}{Logic} & \multicolumn{3}{c}{Retrieval} & Overall \\
Model
& \rot{Sum} & \rot{Sum-C} & \rot{Total}
& \rot{ArgMax} & \rot{ArgMax-R} & \rot{Diff} & \rot{Eq}
& \rot{Lookup} & \rot{Lookup-L} & \rot{Kth}
& \rot{Overall} \\
\midrule
Q25-7B  & \ScoreDelta{90.2}{35.1} & \ScoreDelta{36.8}{0.0}  & \ScoreDelta{89.9}{6.5}  & \ScoreDelta{45.6}{2.1}  & \ScoreDelta{17.7}{-1.0} & \ScoreDelta{16.7}{7.1}  & \ScoreDelta{88.2}{-0.4} & \ScoreDelta{92.7}{-0.3} & \ScoreDelta{48.7}{2.1}  & \ScoreDelta{88.7}{1.6}  & \ScoreDelta{66.2}{3.0} \\
Q3-4B   & \ScoreDelta{88.8}{24.4} & \ScoreDelta{86.5}{29.2} & \ScoreDelta{90.9}{-0.2} & \ScoreDelta{62.3}{5.0}  & \ScoreDelta{37.3}{2.6}  & \ScoreDelta{60.1}{42.4} & \ScoreDelta{94.9}{4.7}  & \ScoreDelta{97.2}{-0.1} & \ScoreDelta{67.0}{-0.5} & \ScoreDelta{\second{95.6}}{-0.1} & \ScoreDelta{81.6}{9.5} \\
Q3-8B   & \ScoreDelta{90.7}{23.9} & \ScoreDelta{\best{98.4}}{29.8} & \ScoreDelta{89.3}{1.7}  & \ScoreDelta{67.4}{4.2}  & \ScoreDelta{39.7}{-0.3} & \ScoreDelta{60.1}{41.4} & \ScoreDelta{\best{97.6}}{7.4} & \ScoreDelta{\second{97.6}}{-0.1} & \ScoreDelta{79.0}{-0.6} & \ScoreDelta{94.1}{0.7}  & \ScoreDelta{84.7}{10.2} \\
Q3-30B  & \ScoreDelta{\second{92.7}}{14.2} & \ScoreDelta{93.5}{9.7} & \ScoreDelta{\second{95.7}}{-0.2} & \ScoreDelta{\best{78.7}}{8.4} & \ScoreDelta{\best{70.7}}{1.4} & \ScoreDelta{\best{62.1}}{17.2} & \ScoreDelta{\second{96.9}}{1.2} & \ScoreDelta{96.1}{2.0} & \ScoreDelta{\best{80.1}}{0.0} & \ScoreDelta{\best{96.5}}{0.0} & \ScoreDelta{\second{88.6}}{6.0} \\
Q3-32B  & \ScoreDelta{\best{95.9}}{19.3} & \ScoreDelta{\second{95.7}}{10.8} & \ScoreDelta{\best{96.4}}{2.3} & \ScoreDelta{\second{72.0}}{3.8} & \ScoreDelta{\second{65.3}}{1.3} & \ScoreDelta{\second{61.6}}{6.5} & \ScoreDelta{\best{97.6}}{3.1} & \ScoreDelta{\best{99.2}}{1.9} & \ScoreDelta{\second{79.6}}{0.5} & \ScoreDelta{94.9}{1.1} & \ScoreDelta{\best{89.1}}{5.8} \\
Olm     & \ScoreDelta{90.2}{38.0} & \ScoreDelta{41.1}{5.4}  & \ScoreDelta{87.3}{-0.3} & \ScoreDelta{56.5}{10.1} & \ScoreDelta{22.9}{-1.1} & \ScoreDelta{24.7}{9.5}  & \ScoreDelta{85.2}{-0.3} & \ScoreDelta{87.0}{-0.5} & \ScoreDelta{56.8}{-0.3} & \ScoreDelta{84.8}{1.6} & \ScoreDelta{67.9}{5.2} \\
NN-s    & \ScoreDelta{51.7}{22.9} & \ScoreDelta{20.0}{8.1}  & \ScoreDelta{42.8}{-1.0} & \ScoreDelta{40.5}{-0.9} & \ScoreDelta{18.5}{-1.5} & \ScoreDelta{15.7}{12.7} & \ScoreDelta{36.1}{-0.4} & \ScoreDelta{67.4}{-1.4} & \ScoreDelta{15.3}{-0.9} & \ScoreDelta{60.2}{-0.7} & \ScoreDelta{39.7}{0.3} \\
NN2-3B  & \ScoreDelta{60.5}{35.6} & \ScoreDelta{19.5}{3.8}  & \ScoreDelta{24.7}{-0.7} & \ScoreDelta{24.9}{-1.0} & \ScoreDelta{5.1}{-0.2}  & \ScoreDelta{32.8}{27.7} & \ScoreDelta{37.9}{-2.1} & \ScoreDelta{55.0}{-2.0} & \ScoreDelta{7.4}{-0.5}  & \ScoreDelta{51.0}{-2.1} & \ScoreDelta{35.1}{2.3} \\
Idef3   & \ScoreDelta{91.2}{53.2} & \ScoreDelta{73.5}{51.3} & \ScoreDelta{45.6}{-1.7} & \ScoreDelta{48.5}{-0.5} & \ScoreDelta{20.5}{-0.8} & \ScoreDelta{\second{61.6}}{56.0} & \ScoreDelta{87.8}{9.0} & \ScoreDelta{86.8}{-0.3} & \ScoreDelta{19.8}{-0.1} & \ScoreDelta{85.2}{2.4} & \ScoreDelta{68.8}{15.4} \\
\bottomrule
\end{tabular}
\endgroup
\end{table*}
Table~\ref{tab:hvs_html_rooter} reports the Table Router Pipeline results when the input is annotated HTML, so performance mainly reflects routing and program planning rather than table reading. Relative to direct QA on the same HTML, gains concentrate on arithmetic-heavy operations, with large improvements in \texttt{Sum} (roughly +14 to +53 points depending on the model) and especially \texttt{Diff} (e.g., +42.4 for Q3-4B and +41.4 for Q3-8B). In contrast, retrieval categories are already near-saturated in the oracle setting and therefore exhibit smaller net changes. Overall, the strongest models are Q3-32B (89.1\%) and Q3-30B (88.6\%), with Q3-8B also competitive (84.7\%); aggregation is handled reliably across them (\texttt{Sum} and \texttt{Total} near or above 90\%, and \texttt{Sum-C} reaching 98.4\% for Q3-8B and 95.7\% for Q3-32B).

\section{Conclusion and Future Work}
We introduced DenTab, a real-world dataset of 2{,}000 dental estimate table images with high-quality HTML annotations, and derived a logic-aware Table QA benchmark with 2{,}208 questions spanning retrieval, aggregation, and consistency reasoning. Our evaluation shows that modern systems can achieve strong TSR performance (up to 92.70 S-TEDS), while TableVQA remains substantially harder (best overall 73.2\%), especially on consistency-difference questions (best 48.5\%). The oracle-HTML analysis further confirms that both perception (table reading) and reasoning contribute to the remaining errors.

We also evaluated a program-assisted TableVQA pipeline that routes arithmetic-heavy questions to deterministic execution, improving results when operating on both table images and annotated HTML. However, DenTab is currently single-domain, and the field still lacks a large \emph{real} multi-domain dataset with joint HTML structure+content annotations, realistic capture artifacts, and industrially grounded reasoning questions. In addition, our pipeline is more expensive than direct VQA because it requires TSR/serialization before program execution (see Appendix for inference time).

Future work will expand DenTab across domains and reduce inference cost via more selective routing and verification, lightweight/distilled TSR, and stronger constraint-based table normalization and validation (e.g., explicit subtotal/total consistency constraints) to improve robustness and generality.

\bibliographystyle{plain} 
\bibliography{references} 

\appendix
\section{Complementary results}
In this section report differents results and additional experiments, we did not include in the main paper due to lack of space.
\subsection{Impact of Image Degradations}
\begin{figure}[t]
\centering
\includegraphics[width=0.95\linewidth]{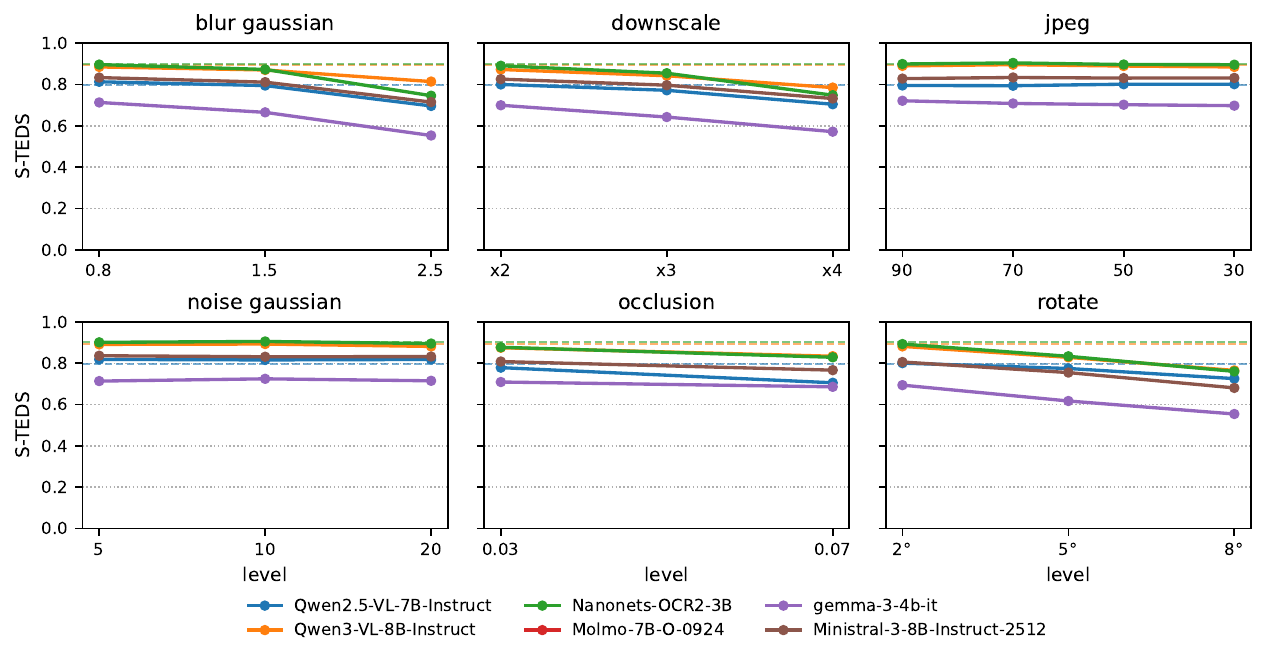}
\caption{S-TEDS under different image degradations; dashed lines indicate each model's clean baseline.}
\label{fig:degrad_panels}
\end{figure}

To quantify robustness to realistic acquisition artifacts, we re-evaluate selected strong TSR systems under controlled degradations (Gaussian blur, downscaling, JPEG compression, Gaussian noise, rotation, and occlusion). Figure~\ref{fig:degrad_panels} reports mean S-TEDS per condition (dashed line: clean baseline). The most damaging degradations are those that reduce effective resolution or distort table geometry. Strong blur and aggressive downscaling yield the largest drops across models (e.g., blur Gaussian 2.5: 69.67\% / 81.47\% / 74.62\% / 55.34\% / 71.54\%; downscale 4: 70.49\% / 78.56\% / 74.87\% / 57.21\% / 73.26\% for Qwen2.5-VL-7B / Qwen3-VL-8B / Nanonets-OCR2-3B / Gemma-3-4B / Ministral-3-8B). Rotation is also harmful at larger angles (rotate 8: 72.57\% / 76.54\% / 76.07\% / 55.43\% / 68.08\%). Occlusion further reduces S-TEDS, especially at higher coverage (occlusion 0.07: 70.55\% / 83.42\% / 82.91\% / 68.60\% / 76.67\%). In contrast, JPEG compression has limited effect (JPEG 30--90 stays near the clean baseline), and Gaussian noise is largely tolerated (typically within a few points of the best noisy setting).

\subsection{Pipeline Accuracy and Throughput}
Table~\ref{tab:pipeline_summary} reports end-to-end pipeline statistics that complement the main accuracy tables, focusing on routing quality and runtime cost. We include (i) pipeline exact match (EM$_{\text{pipe}}$) alongside the direct QA baseline (EM$_{\text{base}}$) for reference, (ii) routing correctness (RouteAcc), i.e., the fraction of questions for which the router selects the ground-truth category, and (iii) efficiency metrics capturing the compute overhead and achieved throughput.

Overhead$_{\text{E2E}}$ is computed as the ratio between the total wall-clock runtime of the full router pipeline (including TSR/HTML parsing, routing, program planning, DSL execution, and any repairs) and the total runtime of direct QA on the same set of questions. QPS$_{\text{pipe}}$ is the achieved end-to-end throughput of the pipeline in questions per second, computed as the number of evaluated questions divided by the pipeline total runtime. Together, these metrics in Table~\ref{tab:pipeline_summary} characterize the practical latency--throughput implications of deploying the router pipeline across model scales, independent of the main paper’s accuracy-focused discussion.

\begin{table*}[t]
\centering
\begingroup
\QATableSetup
\caption{Pipeline accuracy and efficiency summary. EM$_{\text{pipe}}$ and EM$_{\text{base}}$ are overall scores reported from Table 3 and 4 from the paper. RouteAcc is router category accuracy. Overhead$_{\text{E2E}}$ is the end-to-end runtime multiplier relative to direct QA (lower is faster). QPS$_{\text{pipe}}$ is pipeline throughput in questions per second (higher is better).}
\label{tab:pipeline_summary}
\begin{tabular}{l|ccccc}
\toprule
Model & EM$_{\text{pipe}}$ & EM$_{\text{base}}$ & RouteAcc & Overhead$_{\text{E2E}}$ & QPS$_{\text{pipe}}$ \\
\midrule
Q3-32B  & 79.8 & 73.2 & 93.8 & 1.860 & 0.260 \\
Q3-8B   & 73.2 & 61.7 & 83.1 & 1.784 & 0.727 \\
Q3-30B  & 76.0 & 66.2 & 92.2 & 1.699 & 0.272 \\
Q3-4B   & 69.4 & 55.8 & 85.8 & 1.954 & 0.506 \\
Q25-7B  & 58.4 & 56.0 & 73.2 & 1.108 & 1.163 \\
Olm     & 59.6 & 54.5 & 73.6 & 1.282 & 1.057 \\
NN-s    & 35.2 & 29.9 & 24.9 & 1.217 & 0.840 \\
NN2-3B  & 28.8 & 27.5 & 50.0 & 1.351 & 0.797 \\
Idef3   & 31.6 & 30.1 & 82.3 & 2.199 & 0.871 \\
\bottomrule
\end{tabular}
\endgroup
\end{table*}

\begin{figure}[t]
\centering
\begin{minipage}[t]{0.49\linewidth}
  \centering
  \includegraphics[width=\linewidth]{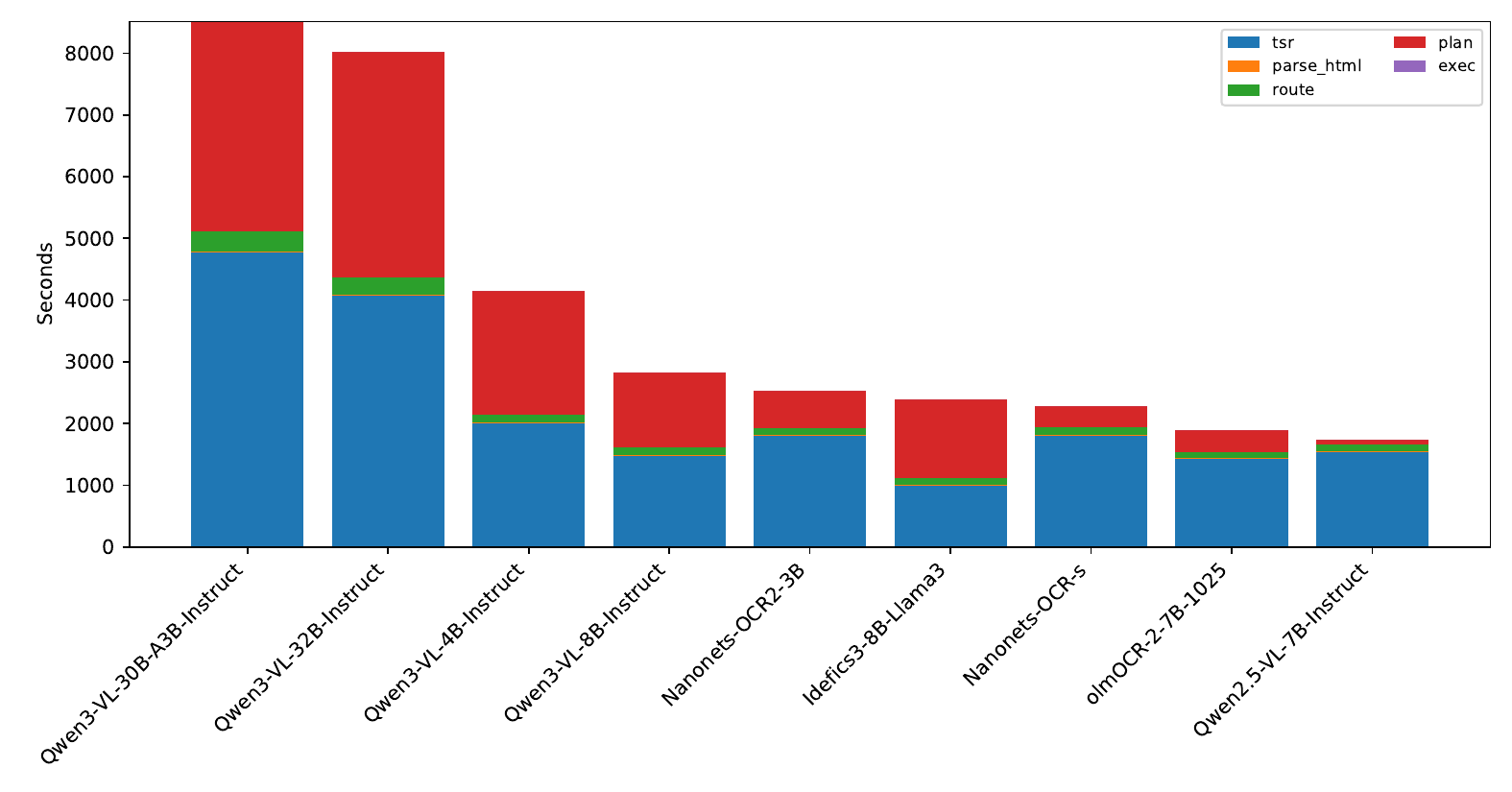}
  \caption{\textbf{(a)} Runtime breakdown.}
\end{minipage}\hfill
\begin{minipage}[t]{0.49\linewidth}
  \centering
  \includegraphics[width=\linewidth]{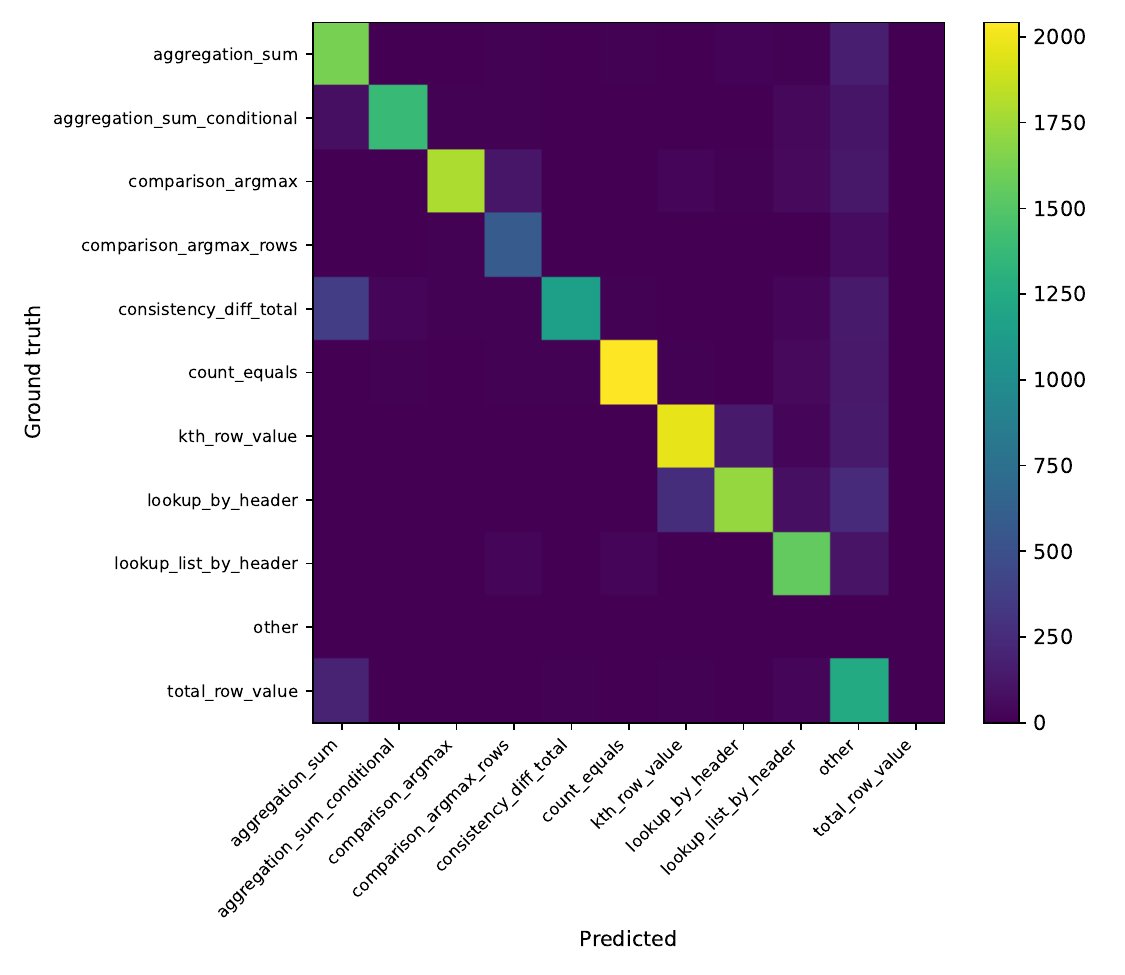}
  \caption{\textbf{(b)} Router confusion matrix.}
\end{minipage}

\caption{Pipeline diagnostics. \textbf{(a)} End-to-end runtime breakdown of the Table Router Pipeline (fractions of wall-clock time per stage: TSR/serialization, routing+program planning, execution, optional repair; baseline direct QA is excluded). \textbf{(b)} Aggregated router confusion matrix over TableVQA categories using the predictions of all the models (rows: ground truth, columns: predicted; normalized per row).}
\label{fig:pipeline_diagnostics}
\end{figure}

\paragraph{Where time is spent.}
Figure~\ref{fig:pipeline_diagnostics} decomposes the end-to-end wall-clock runtime of the router pipeline into its main stages: direct QA (baseline answer), TSR/serialization (HTML generation + parsing), routing and program planning (DSL generation), execution (deterministic evaluation), and optional repair. Across models, TSR/serialization and program planning account for most of the additional cost, while deterministic execution is negligible compared to model inference. This breakdown explains the overhead factors in Table~\ref{tab:pipeline_summary} and highlights where optimization efforts are most impactful (e.g., selective routing, caching of TSR outputs, or lighter TSR backbones).

\paragraph{Router behavior and error modes.}
Figure~\ref{fig:pipeline_diagnostics} reports the aggregated confusion matrix of the router over the TableVQA taxonomy. Most confusions occur between semantically close categories, in particular between arithmetic-heavy classes (e.g., \texttt{aggregation\_sum} vs.\ \texttt{aggregation\_sum\_conditional}, and \texttt{total\_row\_value} vs.\ \texttt{aggregation\_sum}). Retrieval categories are generally more stable, while consistency questions can be misrouted to simpler arithmetic classes when the question mentions multiple quantities. These patterns motivate stricter routing checks for arithmetic categories and justify reporting RouteAcc alongside end-to-end accuracy in Table~\ref{tab:pipeline_summary}.

\section{Prompts, Hyperparameters, and Model Versions} 
\label{sec:supp_prompts}
\paragraph{Prompting}
We use a two-message multimodal chat format with a \texttt{system} instruction and a \texttt{user} message
containing the image and a short reminder. For models whose Hugging Face chat template does not accept the
\texttt{system} role (e.g., templates enforcing strict alternation), we apply an automatic fallback that
merges the system instruction into the first user text while keeping the image unchanged.

\noindent\textbf{System prompt (TSR)}
\small
\begin{MyVerbatim}
You are a vision model that performs TABLE STRUCTURE RECOGNITION (TSR).
Given a document image, you must locate the main tabular region and convert it
into a clean HTML table that encodes the structure of rows, columns, and merged cells.

STRICT OUTPUT FORMAT:
- Output EXACTLY one HTML fragment: a single <table>...</table> element.
- Allowed tags: <table>, <thead>, <tbody>, <tr>, <td>.
- Allowed attributes: colspan, rowspan (positive integers).
- Forbidden: <th>, <caption>, <b>, <i>, <strong>, <em>, <span>, <div>, CSS classes,
  styles, IDs, HTML comments, or any text outside the <table> element.
- DO NOT wrap the table in <html> or <body>.
- DO NOT use ``` fences or language names (like ```html).

STRUCTURAL RULES:
- If the first visual row is a single cell spanning the entire table width (a title/banner),
  IGNORE that row and start from the true column header row.
- Put header row(s) inside <thead>; put all remaining rows inside <tbody>.
- Use one <td> per cell; use colspan/rowspan to represent merged cells so that the grid
  matches the visual layout of the table.
- Preserve the original row and column order. Do NOT reorder columns or rows.
- If there is clearly NO table, output: <table></table>
- If there is multiple tables, output only the first one.

CONTENT RULES:
- Inside each <td>, transcribe the cell text as printed (including accents and punctuation).
- Trim leading and trailing spaces inside cells.
- Keep empty cells as <td></td> when they visually exist, to preserve structure.
- Never invent rows or columns not supported by the image.
\end{MyVerbatim}

\noindent\textbf{User prompt (TSR)}
\small
\begin{MyVerbatim}
Identify the main table in the image and return ONLY its structure and cell contents
as a single HTML <table> that follows the strict rules given above.
\end{MyVerbatim}

\noindent\textbf{System prompt (TableVQA)}
\small
\begin{MyVerbatim}
You answer questions about a table in a document image.
The image contains a main tabular region. You must read the table and answer
each question precisely.

REQUIREMENTS:
- Base your answers only on the table in the image.
- Perform any arithmetic or aggregation that is needed.
- Answer concisely: no explanations, no reasoning steps.
- You MUST return a single JSON object with the following schema:
  {"answers": ["answer_for_question_1", "answer_for_question_2", ...]}
- The list must have EXACTLY as many answers as there are questions.
- Do NOT add commentary, natural language sentences, or Markdown fences.
\end{MyVerbatim}

\noindent\textbf{User prompt (TableVQA)}
\small
\begin{MyVerbatim}
You will be asked N questions about the table in the image.
Answer ALL questions and return ONLY valid JSON with this exact schema:
{"answers": ["answer_for_question_1", "answer_for_question_2", "..."]}
The answers array MUST be in the same order as the questions.

Questions:
Q1: ...
Q2: ...
...
QN: ...

Return only a JSON object with an 'answers' array. No extra text,
no Markdown, no explanations.
\end{MyVerbatim}

\noindent\textbf{System prompt (question category classification)}
\small
\begin{MyVerbatim}
You classify table QA questions.
Choose EXACTLY ONE category label per question.
Return ONLY valid JSON with schema:
{"categories": ["label_for_q1", "label_for_q2", ...]}
No extra text.
\end{MyVerbatim}

\noindent\textbf{User prompt (question category classification)}
\small
\begin{MyVerbatim}
Allowed categories:
lookup_by_header, lookup_list_by_header, kth_row_value, aggregation_sum,
aggregation_sum_conditional, comparison_argmax, comparison_argmax_rows,
count_equals, consistency_diff_total, total_row_value, na_from_empty, other

Questions:
Q1: ...
Q2: ...
...
QN: ...

Return ONLY JSON: {"categories": [..]} (same order).
\end{MyVerbatim}

\noindent\textbf{System prompt (DSL program)}
\small
\begin{MyVerbatim}
You write executable programs to answer questions about a table.
You will be given TABLE_JSON (headers + rows with row_role).
Return ONLY valid JSON.

IMPORTANT RULES:
- Use ONLY ops from the DSL list below.
- Use ONLY header strings that appear EXACTLY in TABLE_JSON.headers.
- Put context ops FIRST (EXCLUDE_ROLES / KEEP_ROLES / FILTER_EQ / SORT), then terminal ops.
- For non-total computations, exclude totals: EXCLUDE_ROLES(["total","subtotal"]).
- For TOTAL questions, use KEEP_ROLES(["total","subtotal"]) and then KTH_ROW.
- For ARGMAX returning a column, use return="col:<header>".
- Return ONLY JSON (no markdown, no explanation).

DSL ops:
- EXCLUDE_ROLES: {"op":"EXCLUDE_ROLES","roles":[...]}
- KEEP_ROLES:    {"op":"KEEP_ROLES","roles":[...]}
- FILTER_EQ:     {"op":"FILTER_EQ","col":"<header>","value":"<string>"}
- SORT:          {"op":"SORT","col":"<header>","order":"asc|desc","numeric":true|false}
- LOOKUP:        {"op":"LOOKUP","key_col":"<header>","key_value":"<string>",
"target_col":"<header>","mode":"first|all","empty_to_na":true|false}
- KTH_ROW:       {"op":"KTH_ROW","k":1|"last","target_col":"<header>","data_only":true|false}
- SUM:           {"op":"SUM","col":"<header>"}
- COUNT:         {"op":"COUNT","col":"<header>","value":"<string>"}
- ARGMAX:        {"op":"ARGMAX","col":"<header>","return":"row_index"|"col:<header>","all_ties":true|false}
- DIFF:          {"op":"DIFF","a":{...},"b":{...}}

Return ONLY JSON with schema: {"programs":[{"qid":1,"ops":[...]}, ...]}.
\end{MyVerbatim}

\noindent\textbf{User prompt (DSL program)}
\small
\begin{MyVerbatim}
TABLE_JSON:
{...}

CATEGORY_HINTS:
["...", "...", ...]

Questions:
Q1: ...
Q2: ...
...
QN: ...

Return ONLY JSON: {"programs":[...]} (qid is 1-based).
\end{MyVerbatim}

\noindent\textbf{System prompt (DSL Repair)}
\small
\begin{MyVerbatim}
You repair an invalid DSL program for a table question.
Return ONLY valid JSON: {"qid":<int>,"ops":[...]}
No extra text.
\end{MyVerbatim}

\noindent\textbf{User prompt (DSL Repair)}
\small
\begin{MyVerbatim}
TABLE_JSON:
{...}

QID: 1
CATEGORY: ...
QUESTION: ...
ERROR: ...

Return ONLY JSON: {"qid":..., "ops":[...]}
\end{MyVerbatim}

\paragraph{Decoding and runtime hyperparameters.}
Inference is performed with vLLM using greedy decoding (temperature 0) and a fixed token budget:
\begin{itemize}
  \item Max new tokens (TSR): 4096
  \item Max new tokens (TableVQA): 1024
  \item Router pipeline max new tokens (planner): 1024
  \item Router pipeline max new tokens (repair): 512
  \item Rooter max repair rounds : 1
  \item Temperature: 0.0
  \item Top-p: 1.0
  \item Repetition penalty: 1.1
  \item Batch size: 1 (multimodal stability)
  \item Image scaling: downscale factor = 1.0
  \item Stopping criteria: a small set of model-dependent stop strings including
  \texttt{```}, \texttt{<|im end|>}, \texttt{<|eot id|>}, \texttt{<|end|>}, and \texttt{</s>}.\\
  For some model families (e.g., Qwen/DeepSeek/Dolphin-v2), \texttt{<|endoftext|>} is added.
\end{itemize}

\paragraph{HTML normalization and sanitization.}
To ensure downstream robustness, we apply light post-processing:
(i) we extract the first \texttt{<table>...</table>} segment if additional text is present,
(ii) we remove accidental wrappers (\texttt{<html>}, \texttt{<body>}),
(iii) we replace \texttt{<th>} by \texttt{<td>} and strip unsupported inline tags
(\texttt{b,i,strong,em,span,div,font,center}). This normalization is purely syntactic and
does not alter cell strings beyond whitespace trimming.

\paragraph{Model loading and implementation details.}
We run all models with vLLM \texttt{LLM.chat()} (multimodal messages) and set:
\begin{itemize}
  \item Precision: \texttt{bfloat16} when supported by the GPU, otherwise \texttt{float16}.
  \item trust remote code: enabled for compatibility with vision model wrappers.
  \item gpu memory utilization: 0.90 and \textbf{enforce eager:} enabled for stability.
  \item limit mm per prompt: \{\texttt{image}: 1\}.
  \item Model-specific overrides: for Qwen-VL families (and Dolphin-v2), we set
  \texttt{mm processor kwargs=\{min pixels=28*28, max pixels=1280*28*28\}}.
  For Qwen3-VL models, we set \texttt{max model len=32768}. For DeepSeek-VL2,
  we include a small \texttt{hf overrides} to ensure correct architecture resolution.
\end{itemize}

\paragraph{Evaluated models (Hugging Face identifiers).}
We evaluate the following vision-language models:
\begin{itemize}
  \item \texttt{Qwen/Qwen2.5-VL-7B-Instruct}
  \item \texttt{Qwen/Qwen3-VL-4B-Instruct}
  \item \texttt{Qwen/Qwen3-VL-8B-Instruct}
  \item \texttt{Qwen/Qwen3-VL-30B-A3B-Instruct}
  \item \texttt{Qwen/Qwen3-VL-32B-Instruct}
  \item \texttt{ByteDance/Dolphin-v2}
  \item \texttt{allenai/olmOCR-2-7B-1025}
  \item \texttt{nanonets/Nanonets-OCR-s}
  \item \texttt{nanonets/Nanonets-OCR2-3B}
  \item \texttt{deepseek-ai/deepseek-vl2-small}
  \item \texttt{HuggingFaceM4/Idefics3-8B-Llama3}
  \item \texttt{google/gemma-3-4b-it}
  \item \texttt{google/gemma-3-12b-it}
  \item \texttt{google/gemma-3-27b-it}
  \item \texttt{mistralai/Ministral-3-8B-Instruct-2512}
  \item \texttt{mistralai/Ministral-3-14B-Instruct-2512}
\end{itemize}

\section{Dataset annotation}
\begin{figure}[t]
\centering
  \includegraphics[width=\linewidth]{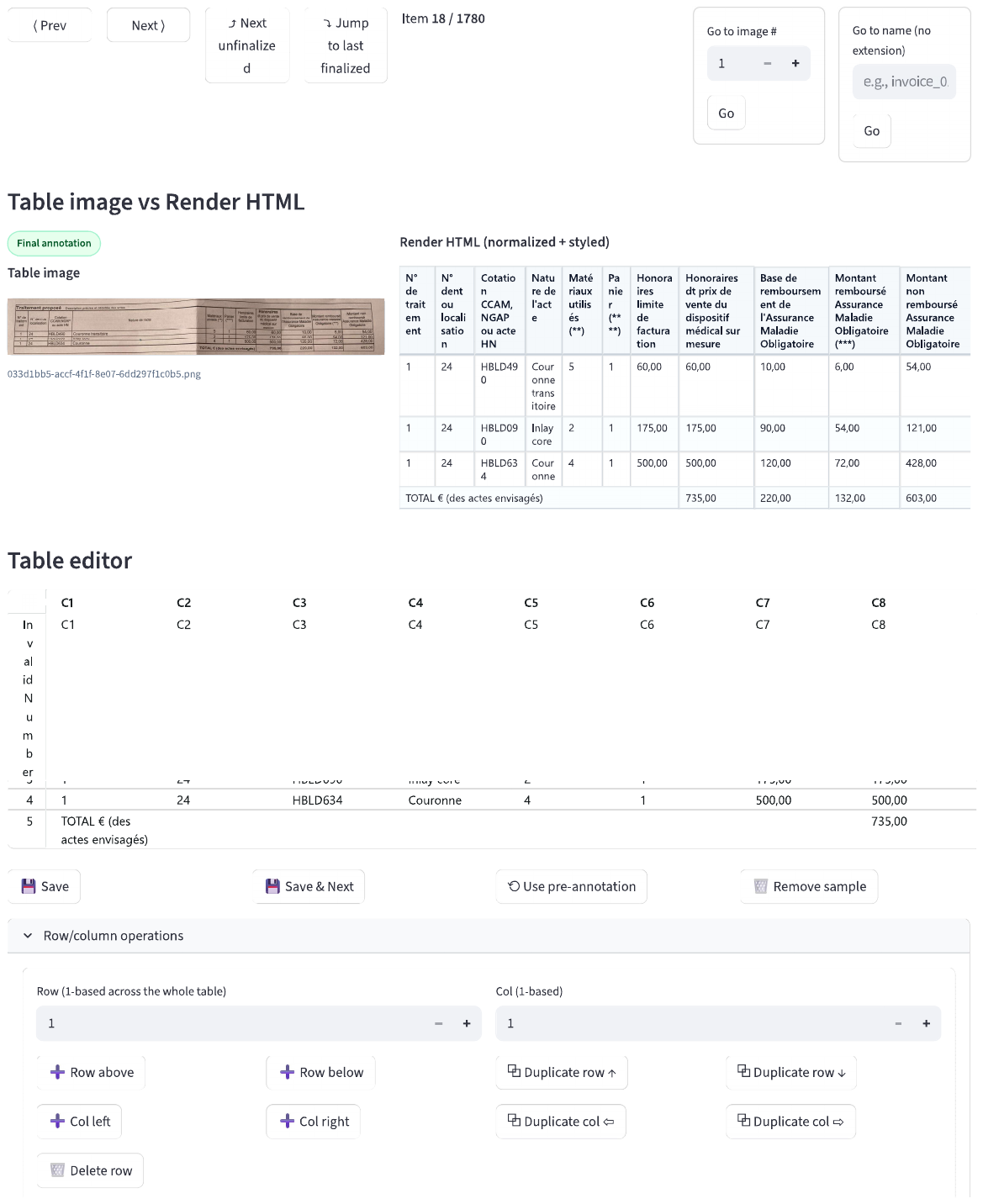}
  \caption{Web UI of our built in HTML annotator}
\label{fig:webapp}
\end{figure}

To annotate the dataset we have built an annotator web application using streamlit, it enable to load pre-annotations and then modify the cells content, and allows multiple operations like row operations (deletion, adding, duplicating), cells spanning operations and headers operations. It make correcting the structure and content errors more easy as we can see the results directely in the renderer, as you can see in the Figure~\ref{fig:webapp}.

\section{Table Recognition samples}
\begin{figure}[t]
\centering
  \includegraphics[width=\linewidth]{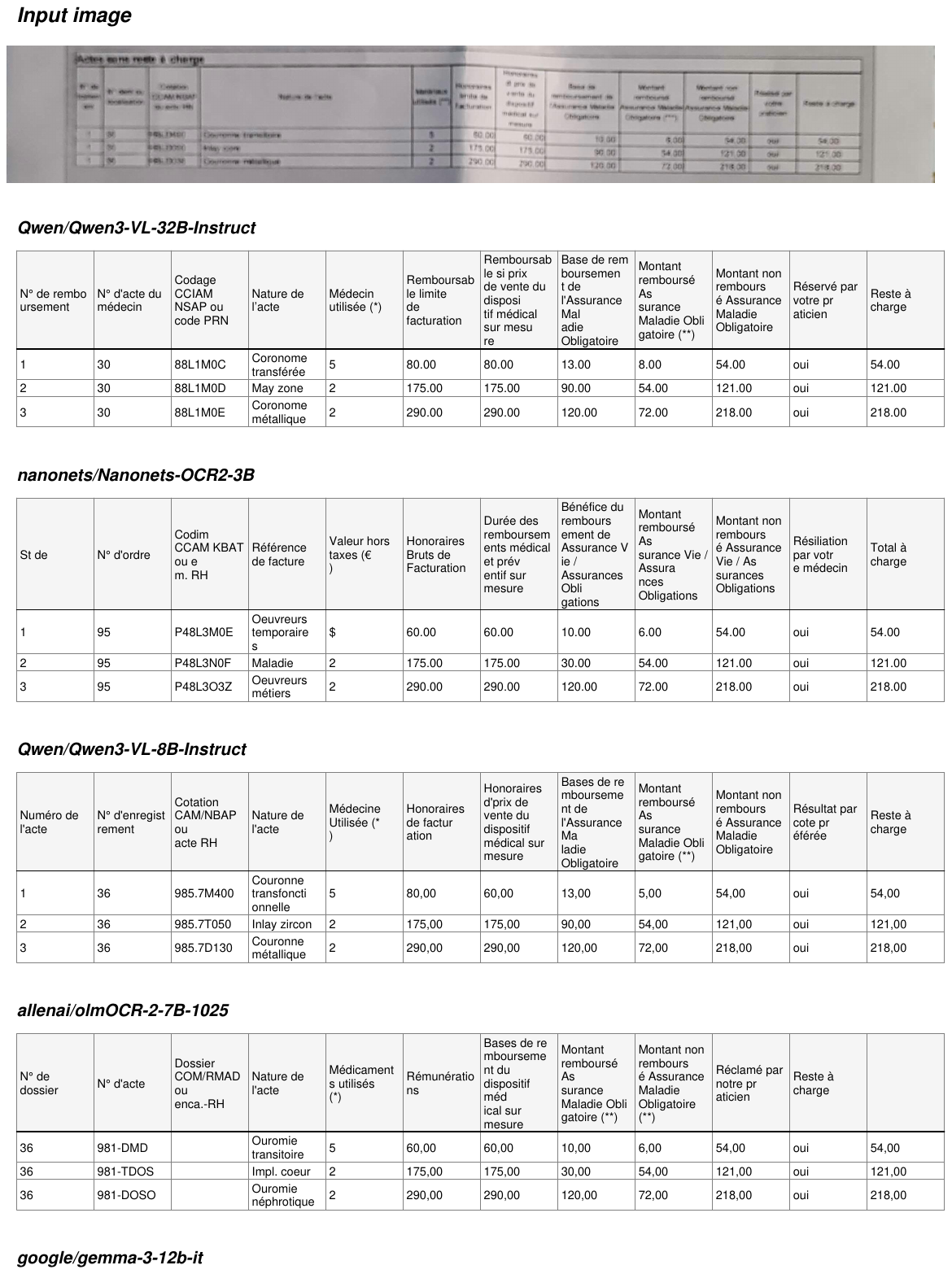}
  \caption{Input image with VLMs output rendered in HTML (sample1)}
\label{fig:tr1}
\end{figure}

\begin{figure}[t]
\centering
  \includegraphics[width=\linewidth]{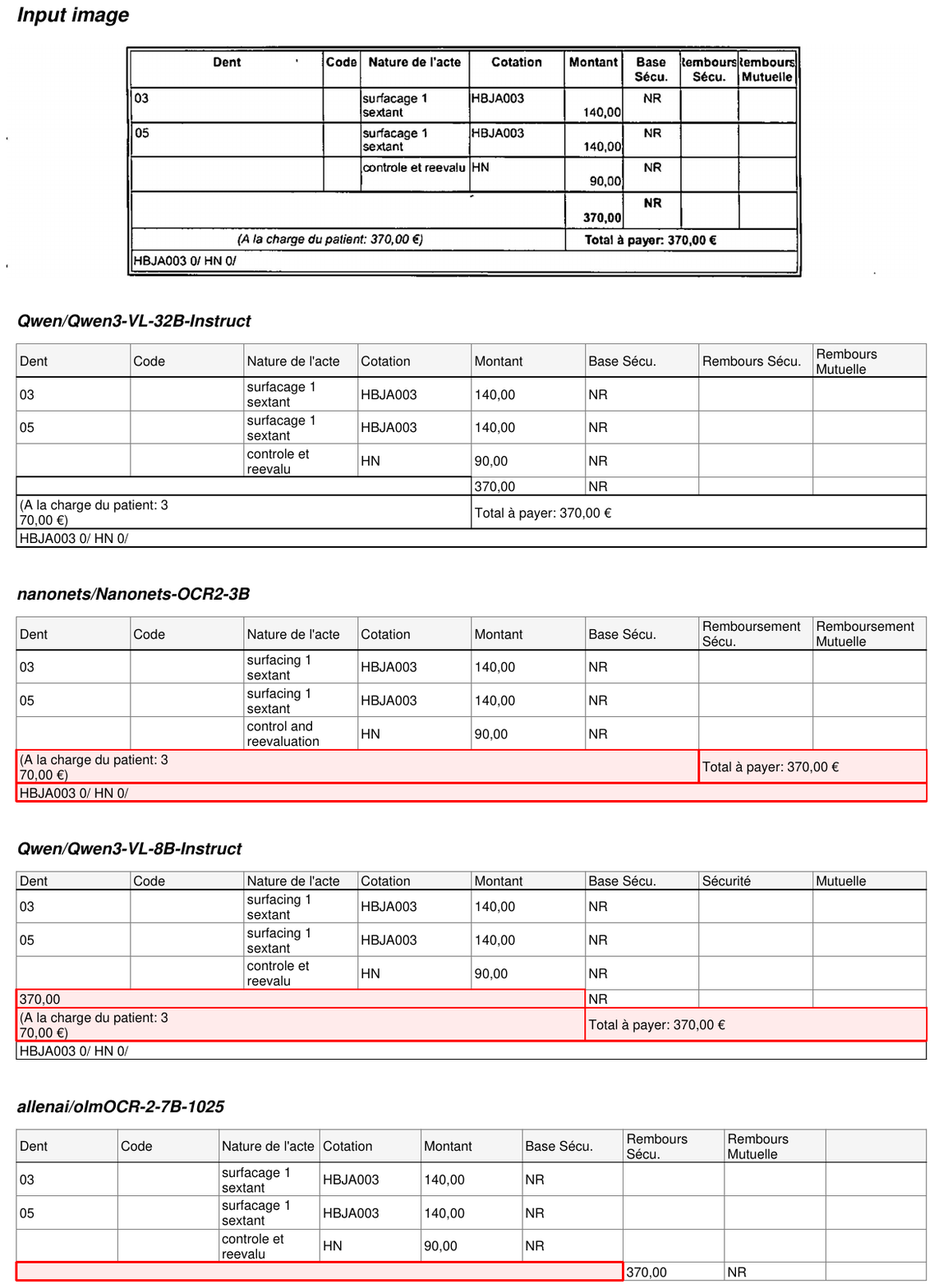}
  \caption{Input image with VLMs output rendered in HTML (sample2)}
\label{fig:tr2}
\end{figure}

See bellow VLMs representatives predictions rendered with HTML to facilitate the comparison and detect the structural errors. See figure~\ref{fig:tr1} and~\ref{fig:tr2}.

\section{Direct TableVQA}
\begin{figure}[t]
\centering
  \includegraphics[width=\linewidth]{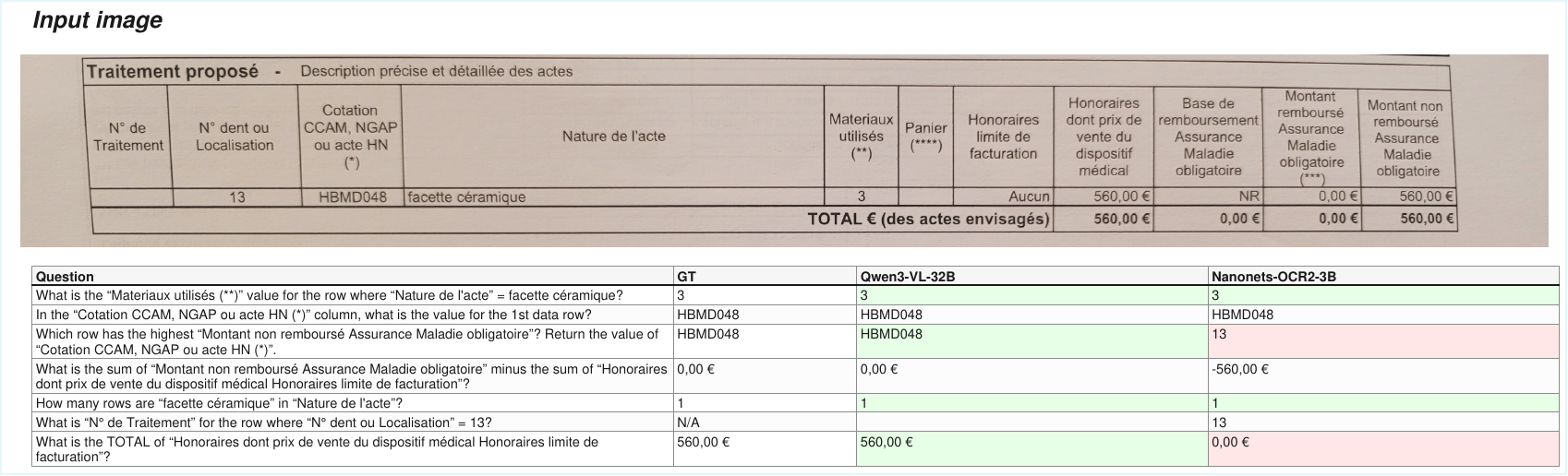}
  \caption{Sample 1}
\label{fig:tqa1}
\end{figure}

\begin{figure}[t]
\centering
  \includegraphics[width=\linewidth]{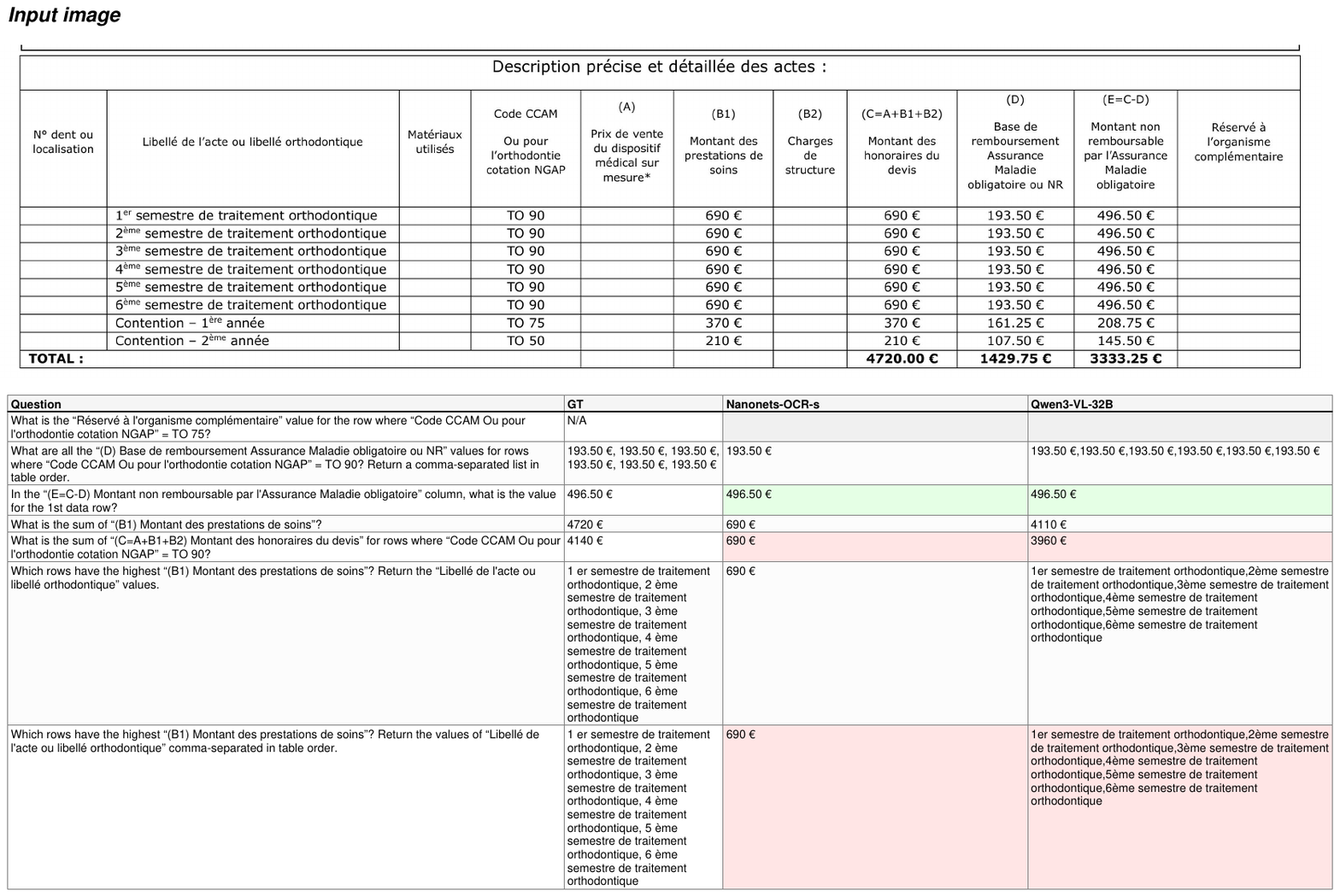}
  \caption{Sample 2}
\label{fig:tqa2}
\end{figure}

See bellow VLMs representatives predictions on direct TableVQA. See figure~\ref{fig:tqa1} and~\ref{fig:tqa2}.

\end{document}